\documentclass[journal]{IEEEtran}

\usepackage{amsmath}
\usepackage{amssymb}
\usepackage{graphicx}
\usepackage{cite}
\usepackage{algorithm}
\usepackage{algorithmic}
\usepackage{multirow}
\usepackage{subfigure}
\usepackage{color}
\usepackage{bm}
\usepackage{fancyhdr} %
\usepackage{booktabs}

\usepackage{ulem}

\ifCLASSINFOpdf
\else
\fi

\usepackage{array}

\ifCLASSOPTIONcompsoc
 \usepackage[caption=false,font=normalsize,labelfont=sf,textfont=sf]{subfig}
\else
 \usepackage[caption=false,font=footnotesize]{subfig}
\fi

\begin{document}

\title{TSGB: Target-Selective Gradient Backprop for Probing CNN Visual Saliency}

\author{
        Lin~Cheng,
        Pengfei~Fang,
        Yanjie~Liang,
        Liao~Zhang,
        Chunhua~Shen,
        Hanzi~Wang$^*$,~\IEEEmembership{Senior~Member,~IEEE}

\thanks{This work is supported by the National Natural Science Foundation of China (Grant No.\  U21A20514 and 61872307).}        
 \thanks{L. Cheng, L. Zhang, and H. Wang are with Fujian Key Laboratory of Sensing and Computing for Smart City, School of Informatics, Xiamen University, Xiamen 361005, 
 China (e-mail:
 cheng.charm.lin@hotmail.com, 
 leochang@stu.xmu.edu.cn, 
 hanzi.wang@xmu.edu.cn).}
 \thanks{P. Fang is with College of Engineering and Computer Science, 
 Australian National University, Canberra, ACT 2601, Australia (e-mail: Pengfei.Fang@anu.edu.au).}
 \thanks{Y. Liang is with the Peng Cheng Laboratory, Shenzhen 518000, P.R. China (e-mail: liangyj@pcl.ac.cn).}
 \thanks{C. Shen is with
Zhejiang University, China (e-mail: Chunhua@me.com).
}
 \thanks{*Corresponding author}
}

\maketitle

\thispagestyle{fancy} 
\lhead{} 
\chead{} 
\rhead{} 
\lfoot{} 
\cfoot{\thepage} 
\rfoot{} 
\renewcommand{\headrulewidth}{0pt} 
\renewcommand{\footrulewidth}{0pt} 

\pagestyle{fancy}
\cfoot{\thepage}

\begin{abstract}
The explanation for deep neural networks has drawn extensive attention
in the deep learning community over the past few years.
In this work, we study the visual saliency, a.k.a.\ 
visual explanation, to interpret convolutional neural networks. 
Compared to iteration based saliency methods, single backward pass based saliency methods benefit from faster 
speed, and they are widely used in downstream visual tasks. 
Thus, we focus on single backward pass based methods.
However, existing methods in this category
struggle to successfully produce fine-grained saliency maps concentrating on specific target classes. That said, 
producing faithful saliency maps satisfying both target-selectiveness and fine-grainedness using a single backward pass 
is a challenging problem in the field.
To mitigate this problem, we revisit the gradient flow inside the network, and find that the entangled semantics and original weights may disturb the propagation of target-relevant saliency. 
Inspired by those observations, we propose a novel visual saliency method, termed
\textit{Target-Selective Gradient Backprop} (TSGB), which leverages rectification operations to effectively emphasize target classes and further efficiently propagate the saliency to the image space, thereby generating \textit{target-selective} and \textit{fine-grained} saliency maps. 
The proposed TSGB consists 
of two components, namely, TSGB-Conv and TSGB-FC, which rectify the gradients for convolutional layers and fully-connected layers, respectively. 
Extensive qualitative and quantitative experiments on the ImageNet and Pascal VOC datasets show that the proposed method achieves more accurate and reliable results than the other competitive methods.
Code is available at https://github.com/123fxdx/CNNvisualizationTSGB.

\end{abstract}

\begin{IEEEkeywords}
Model interpretability, explanation, 
saliency map,
CNN visualization
\end{IEEEkeywords}

\section{Introduction}
\label{sec:intro}
\IEEEPARstart{I}{n} recent years, deep convolutional neural networks (CNNs) have revolutionized various computer vision tasks, including object classification~\cite{tip18cls,tip21cls}, semantic segmentation~\cite{tip20seg,tip20Cseg}, low-level image processing~\cite{tip16dehaze}, etc. 
However, human's knowledge on how deep models make decisions is still limited, which affects the trustworthiness of such a “Black Box” in the deep learning community. Moreover, this trustworthiness issue limits the development of real-world applications, e.g., autonomous driving~\cite{iccv_drive} and medical diagnoses~\cite{CVPR_medical}.

To interpret the working mechanism of deep neural networks, some explanation methods~\cite{TIP_fixation,NIPS19concep,aaai19/LinguisticExplanations,aaai20/InterpretationsNormalization,zhou2016cvpr} have been developed 
to help humans understand what we can trust and how we can improve the networks. 
This paper studies the visual saliency~\cite{GradBP,mm20}, w.r.t.\  the target classes, to explain how CNNs make decisions for given input images. The visual saliency, a.k.a.\  visualization, or visual explanation, aims to highlight important features, 
which highly contribute to the network predictions. 
In addition, visual saliency is also a useful technique for some downstream tasks, e.g., weakly-supervised
vision~\cite{cvpr17Erasing,cvpr19weaklySeg}, person re-identification
~\cite{reid_cvpr19,aug_reid_cvpr19,Fang_2019_ICCV}, knowledge distillation~\cite{CVPR19learn}, etc.

In general, iteration based methods and single backward pass based methods are two dominant groups of methods to 
probe 
the visual saliency. Iteration based methods can localize the important regions in images by conducting iterative feedforwards or backwards~\cite{deconv,Fong_2017,RISE_petsiuk2018rise,LIME2016kdd,YulongWang20TMM}. Such an approach 
is time-consuming, and it may
introduce  
adversarial noise to saliency maps~\cite{Fong_2017}.
In contrast, single backward pass based methods offer the advantage of being computationally efficient, without introducing adversarial noise. To benefit from these
properties, 
this work focuses on the single backward pass based method to study the visual saliency.

\begin{figure}[!t]\centering
\includegraphics[width=0.9\linewidth]{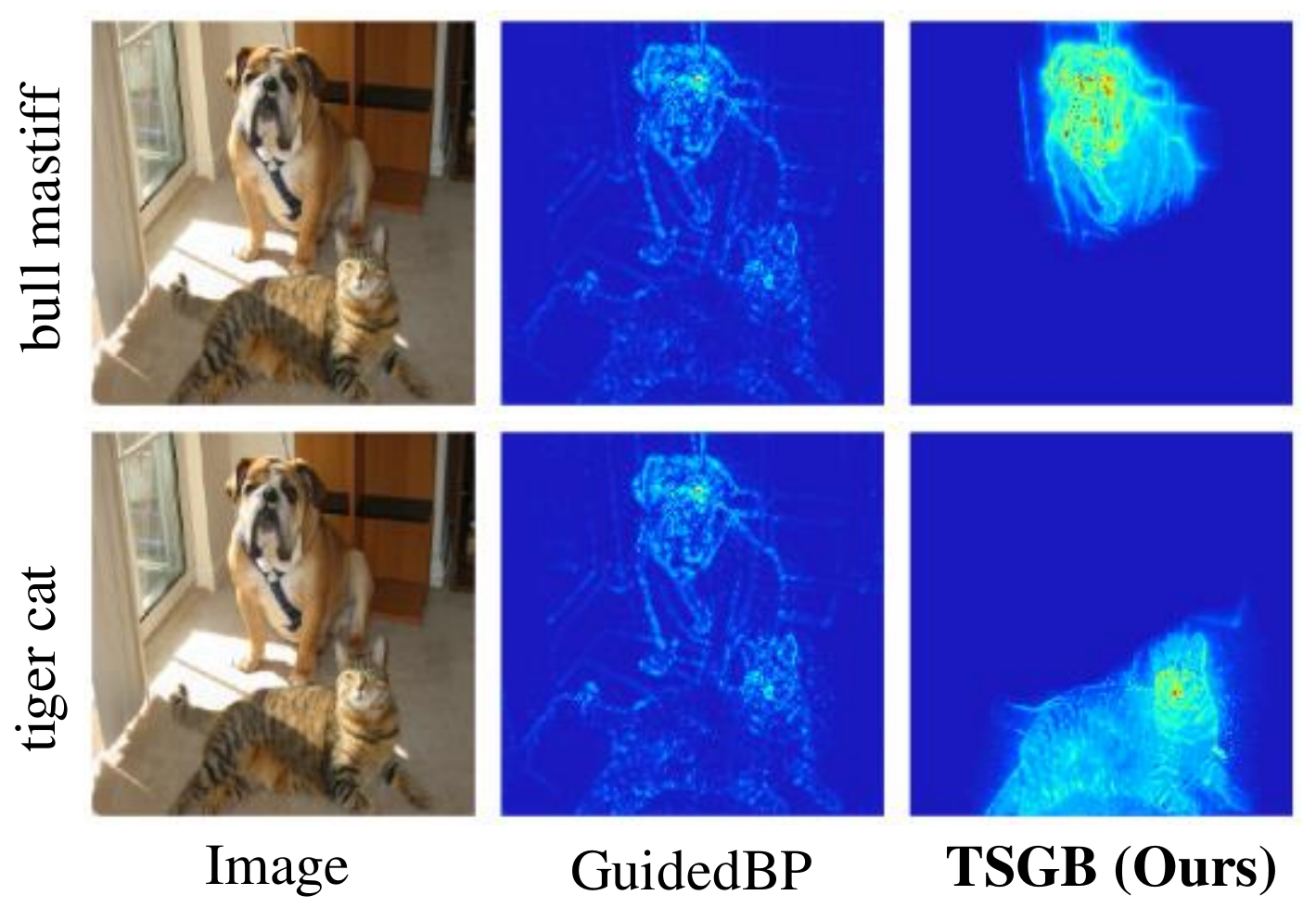}
\caption{\small{Comparison of saliency maps w.r.t.\  target-selectiveness for the predictions of ``bull mastiff'' and ``tiger cat''. GuidedBP~\cite{GradBP} produces two similar saliency maps, while the proposed TSGB produces the discriminative saliency maps.}
}
\label{fig:fist example}
\end{figure}

\begin{figure}[!t]\centering
\includegraphics[width=\linewidth]{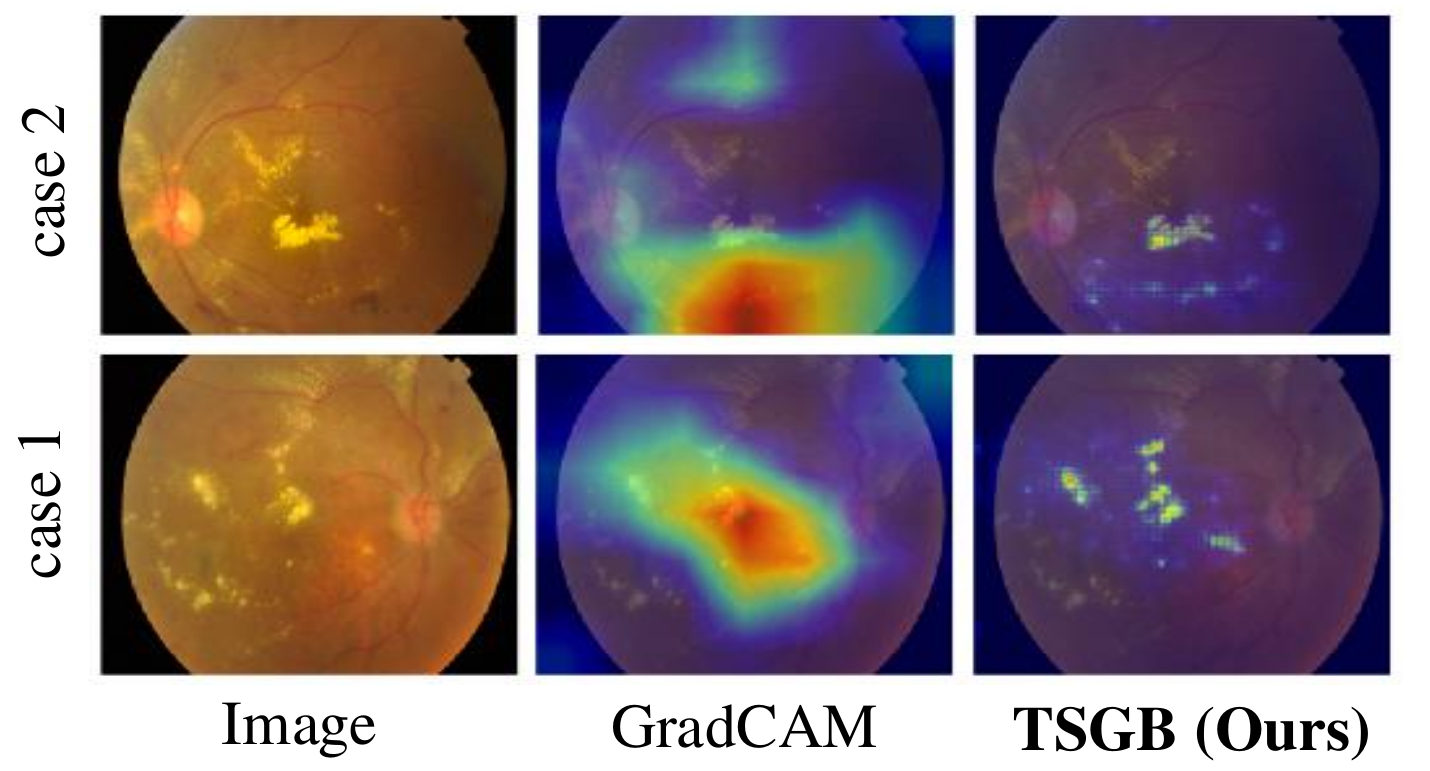}
\caption{\small{Comparison of saliency maps w.r.t. fine-grainedness for the predictions of two cases of diabetic retinopathy. GradCAM~\cite{gradcam} produces two coarse saliency maps, while the proposed TSGB can produce the fine-grained saliency maps.}}
\label{fig:second example}
\end{figure}

Among single backward pass based methods, many
works, e.g., GradBP~\cite{GradBP}, GuidedBP~\cite{GGBP_springenberg2014striving}, and FullGrad~\cite{fullgrad}, have been proposed to exploit the gradient to generate saliency maps, where dominant objects of input images are highlighted. However, such methods often fail to focus on the target class, leading to inferior results w.r.t.  the target category of interest.
As shown in Fig.~\ref{fig:fist example}, GuidedBP produces two similar saliency maps, which cannot focus on the target class. 
Assume an extreme situation: an explanatory result for a selected target turns out target-agnostic, which is meaningless for the explanatory work. 
Other works, e.g., EBP~\cite{ebp} and GradCAM~\cite{gradcam}, attempt to leverage the top-down relevance or the weighted activation maps to produce class-discriminative explanations. However, they
fail to backpropagate the saliency to the input image space, thereby resulting in coarse saliency maps. 
Such coarse explanations are inadequate when fine-grained localization becomes a concern. For example, in the domain of medical image predictions, the fine-grained explanations are essential to discriminate the fine biological tissues~\cite{Wagner_2019_CVPR,mm20}. As shown in Fig.~\ref{fig:second example}, GradCAM produces two coarse saliency maps, which cannot reveal the fine-grained patterns.
Although 
many efforts have
been made to study single backward pass based methods, developing an explanation approach
that satisfies both 
class-level \textit{target-selectiveness} and pixel-level \textit{fine-grainedness}
still needs further investigation.

In this paper, we attempt to address this challenging problem by taking a step back and rethinking the discipline of gradients inside neural networks.
As noticed in the literature~\cite{aaai/ZhangCSWZ18,cvpr/ZhangWZ18a}, the network nodes in the intermediate layers may couple different semantic concepts. 
Interestingly,
we find that even the final hidden layer before the output/prediction layer may contain entangled semantics. 
Such entangled nodes severely affect the target-selectiveness property when propagating the target attribution to the lower layers using gradients. On the other hand, we also observe that the saliency maps can be disturbed by the gradients using pre-trained parameters in convolutional layers. 
This impedes the attribution passing to the bottom to obtain fine-grained saliency maps. 

Inspired by the above observations,
we propose a novel visual saliency method, termed 
\textit{\underline{T}arget-\underline{S}elective \underline{G}radient \underline{B}ackprop} (TSGB), to generate target-specific and fine-grained saliency maps, which can explain how CNNs make decisions. The proposed TSGB consists of two modules, i.e., a target selection module for fully-connected (FC) layers and a fine-grained propagation module for convolutional (Conv.) layers. 
The target selection module exploits the contributions of sub-nodes to the target node, and emphasizes the negative connections by the ratio of positive contributions to negative contributions, which can disentangle the target class from the irrelevant classes and background in features. 
The fine-grained propagation module leverages the ratio of feature responses between two consecutive layers to propagate the visual saliency 
from the feature space
to the image space.

The main contributions of this paper are summarized as follows:

\begin{itemize}
\item We study the influence of entangled semantics and original gradients on the backprop of visual saliency.
Based on our findings, we propose a novel visual saliency method, i.e., TSGB, to explain CNNs' decisions. 
To our best knowledge, this is the first work to 
generate target-selective and fine-grained saliency maps in a single backward pass.

\item We design a target selection module, i.e., TSGB-FC, for the backprop of FC layers. TSGB-FC adaptively enhances the negative connections inside the networks to make the visual saliency effectively focus on the target class.

\item We devise a fine-grained propagation module, i.e., TSGB-Conv, for the backprop of Conv. layers and other advanced layers. TSGB-Conv exploits the information of feature maps rather than model parameters to efficiently produce high-resolution saliency maps.
\end{itemize}

Extensive experiments show the superiority of the proposed TSGB against the competitive methods in target-selectiveness, fine-grainedness, running speed, explanatory generalization, and faithfulness. Moreover, TSGB can be employed to diagnose the biases in the model and dataset. Furthermore, TSGB can be used to help human interpret the CNN model trained for medical diagnoses, and locate the critical biological structures.

The remainder of this paper is organized as follows: In Section \ref{sec:relatedwork}, some related works are described. In Section \ref{sec:Analysis}, the factors disturbing the target-selectiveness and fine-grainedness during gradients backprop are analyzed. Based on the analysis, the proposed method, including the target selection module and the fine-grained propagation module, is presented in Section \ref{sec:method}. In Section \ref{sec:expt}, qualitative  and  quantitative experiments are conducted on various tasks to validate our method against the competitors. Conclusion and discussion are drawn in Section \ref{sec:conclusion}.

\section{Related work}\label{sec:relatedwork}

A variety of saliency methods have been studied to interpret the decisions made by CNNs. Those methods can be categorized into two groups according to the number of processing of feedforward and backward, namely, single backward pass based methods as well as iteration based methods (i.e., multiple feedforward and backward pass based methods). We first focus on discussing three kinds of single backward pass based methods in Section \ref{subsec:A}. We then review several iteration based methods in Section \ref{subsec:B}.

\subsection{Single Backward Pass Based Methods}\label{subsec:A}

\textit{1) Gradient Related Methods:} 
GradBP~\cite{GradBP} is one of the pioneering works for exploring visual saliency, which computes the gradient of the class score w.r.t.\  the input image to visualize the importance heatmap. Thereafter, GuidedBP~\cite{GGBP_springenberg2014striving} and Deconvolution~\cite{deconv} modify the backpropagated gradients, which makes the saliency maps sharper and clearer. 
Note that their explanation results fail to concentrate on the selected target~\cite{eccv/MahendranV16,gradcam}.
As the most recent work, FullGrad~\cite{fullgrad} improves the saliency maps by considering the multi-layer gradients aggregation.

\textit{2) Relevance Related Methods:} Layer-wise Relevance Propagation~\cite{LRP_Bach_2015} and Deep Taylor decomposition ~\cite{Taylor_Montavon_2017} explain the networks by decomposing the contribution of the target layer by layer. These methods pay attention to extensive existing objects, similar to~\cite{GGBP_springenberg2014striving}. Excitation Backprop (EBP)~\cite{ebp} uses the contrastive marginal winning probability to propagate the top-down attention.
DeepLIFT ~\cite{LIFT_shrikumar2017learning} assigns the attribution by comparing the difference between the input and the reference data. CNN Fixation~\cite{TIP_fixation} measures the contributions between a pair of consecutive layers to uncover the pixel coordinates of saliency regions. 

\textit{3) Activation Related Methods:}
CAM ~\cite{zhou2016cvpr} and the generalized version GradCAM~\cite{gradcam} utilize the gradient to weigh the feature maps to localize the important regions. 
This type of method is still the optimal one as noted in~\cite{20cvpr_benchmark}.
Guided GradCAM~\cite{gradcam} ensembles GuidedBP and GradCAM, which actually needs more than one backward pass, and its target-selectiveness almost depends on GradCAM.

Despite that these single backward pass based methods are advanced in visual saliency, their explanatory results cannot satisfy the properties of target-selectiveness and fine-grainedness simultaneously.

\subsection{Iteration Based Methods}\label{subsec:B}
Another type of visual saliency method is based on iterations. 
Perturbation related methods, such as Occlusion~\cite{deconv}, Meaningful Perturbation~\cite{Fong_2017}, RISE~\cite{RISE_petsiuk2018rise} and LIME~\cite{LIME2016kdd},
evaluate the output scores by occluding the input iteratively, which takes much running time and it is possible to introduce adversarial noise.
Most recently, Score-CAM~\cite{Score-CAM} masks the input according to the intermediate activation maps and repeatedly performs feedforwards N times (i.e., the number of activation maps) to obtain the importance scores.
Optimization related methods,
including Feedback~\cite{Cao2015Look}
and FGVis~\cite{Wagner_2019_CVPR}, add the complex switch structure into the network and iteratively optimize the objective function to achieve the saliency maps.
Some integration related methods, such as SmoothGrad~\cite{smilkov2017smoothgrad}, IntegratedGrad~\cite{intg17icml}, and Integrated Grad-CAM~\cite{integ_gradcam}, can be regarded as the 
ensembles
over single propagation methods. We can also take advantages of 
these 
ensembles to improve our method.
These iteration based methods are time-consuming and they do not achieve the optimal performance.

Compared to iteration based methods, single backward pass based methods run faster, 
and they are less likely to introduce
the adversarial noise. Hence, we focus on investigating the single backward pass based visual saliency. 
Unlike all of these methods, our method rectifies the gradient backprop, which satisfies both the target-selectiveness and fine-graininess in a high-speed manner.

\section{Analysis}\label{sec:Analysis}
During the procedure of generating the visual saliency, what exactly affects the selectiveness of the target in saliency maps? What disturbs the visual attribution when backpropagating saliency maps from the top layer to the low layer and what makes the visualized results rough rather than fine-grained? Driven by these two 
crucial
questions, we attempt to explore the problem by revisiting the discipline of gradients inside the networks, as gradients indeed contain inherent and fundamental properties of the networks and they have been employed by many works~\cite{GradBP,GGBP_springenberg2014striving,gradcam,smilkov2017smoothgrad,intg17icml,fullgrad} for explanations.

\begin{figure}[t] 
    \centering
    \subfigure[Test for ``dog head'']{
    \includegraphics[width=3.4cm]{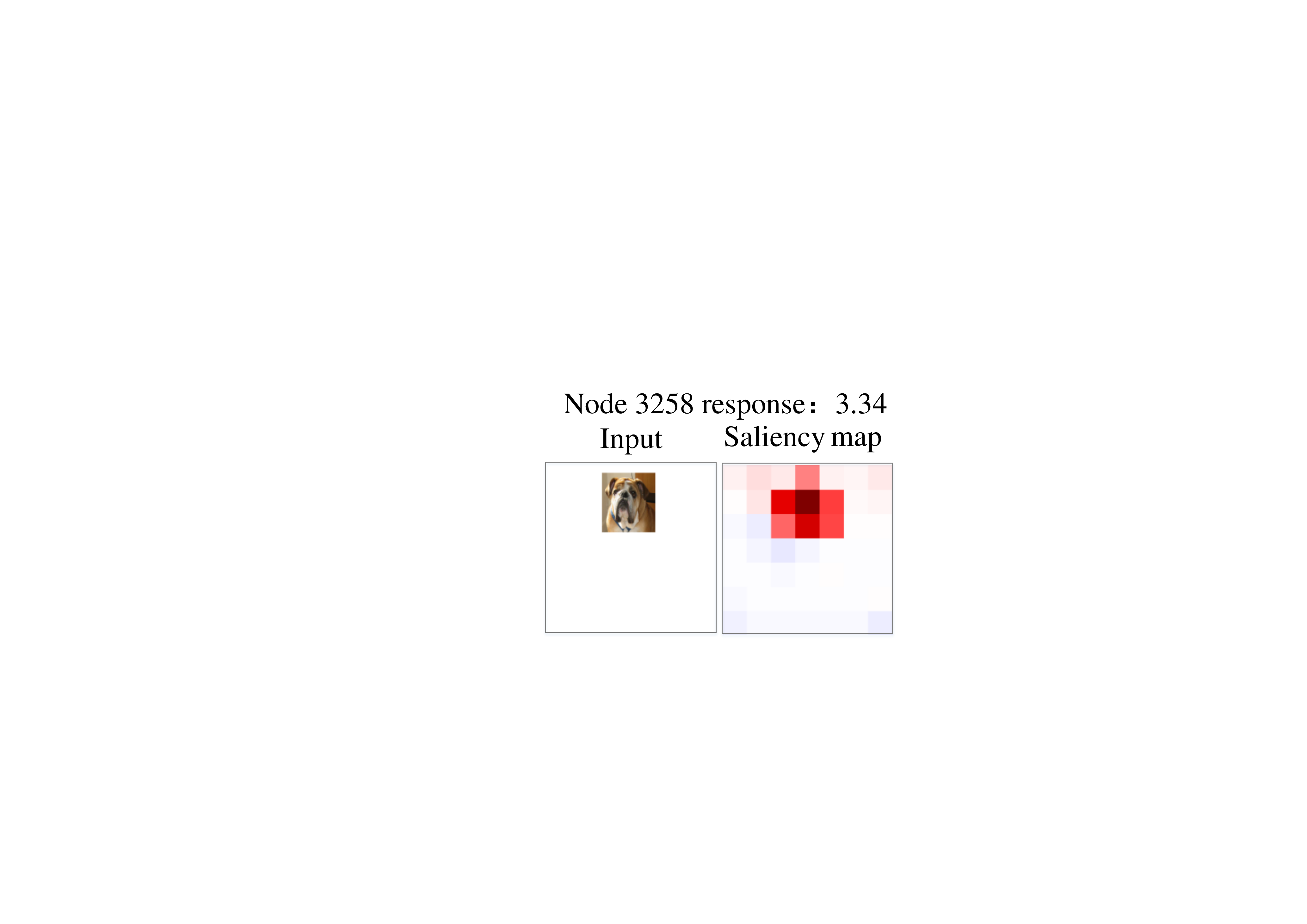}
    } \hspace{-2.8mm}
    \subfigure[Test for ``cat head'']{
    \includegraphics[width=3.4cm]{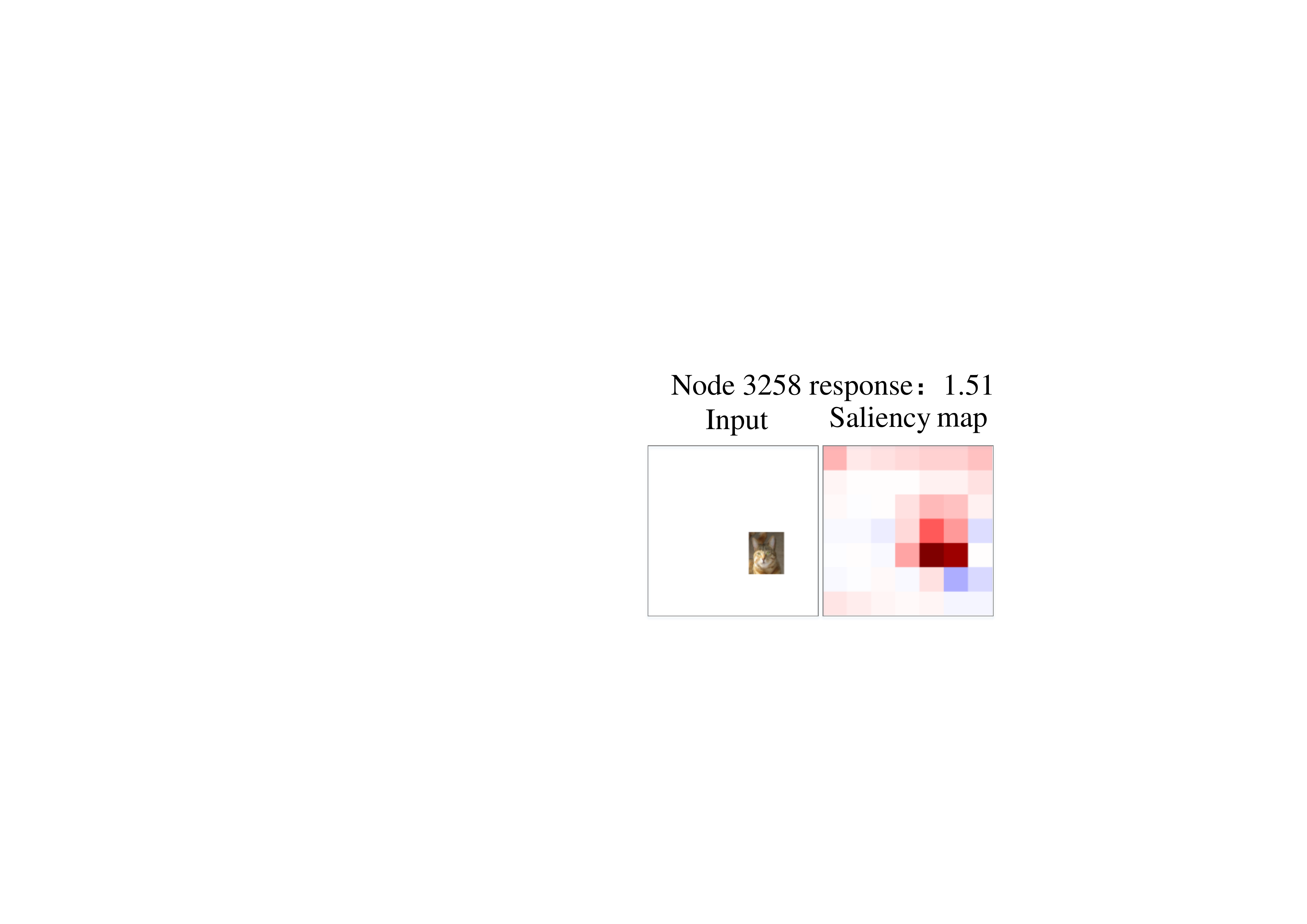}
    } \hspace{-2.8mm}
    \subfigure[Test for ``background'']{
    \includegraphics[width=3.4cm]{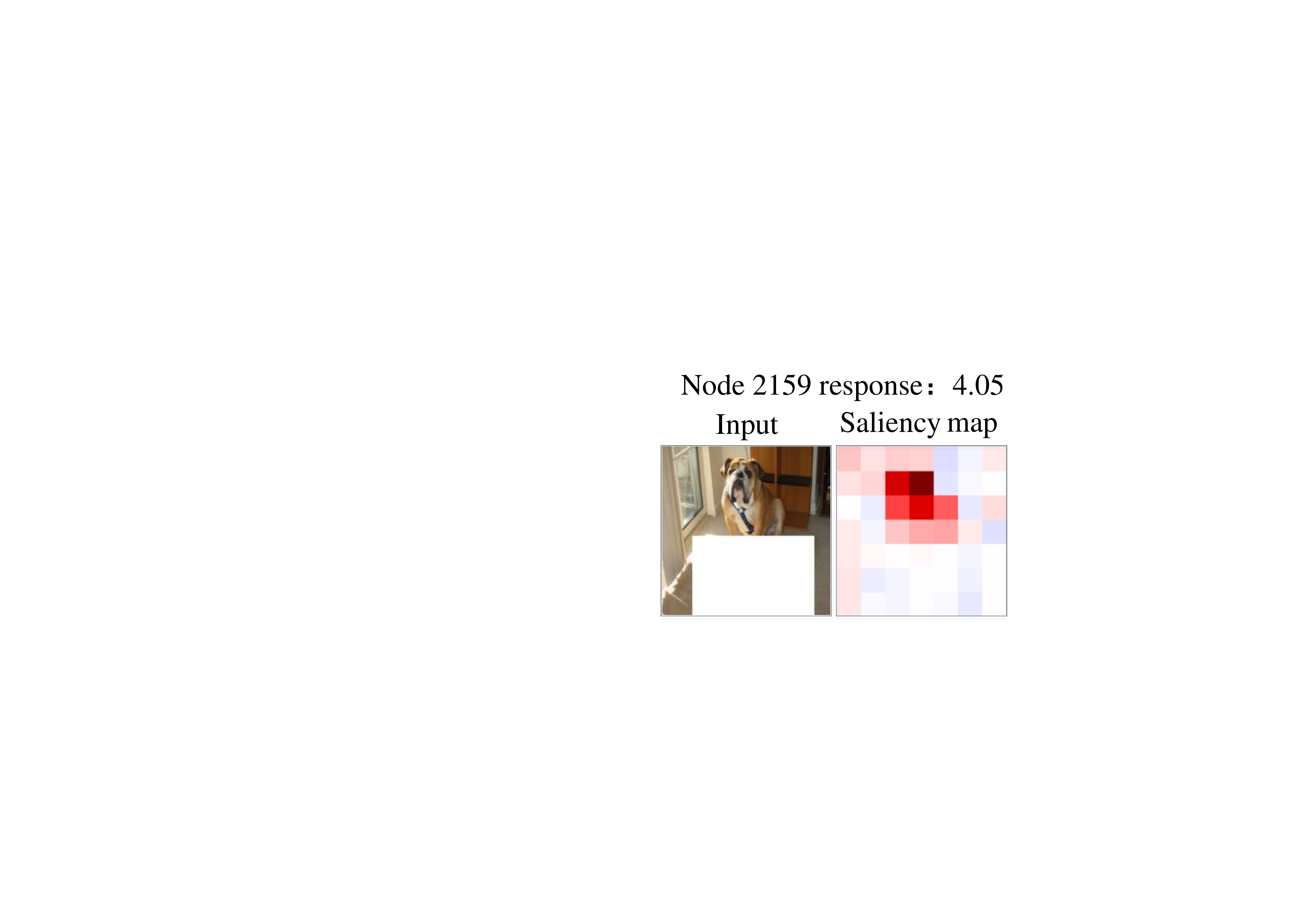}
    } \hspace{-2.8mm}
    \subfigure[Test for ``cat'']{
    \includegraphics[width=3.4cm]{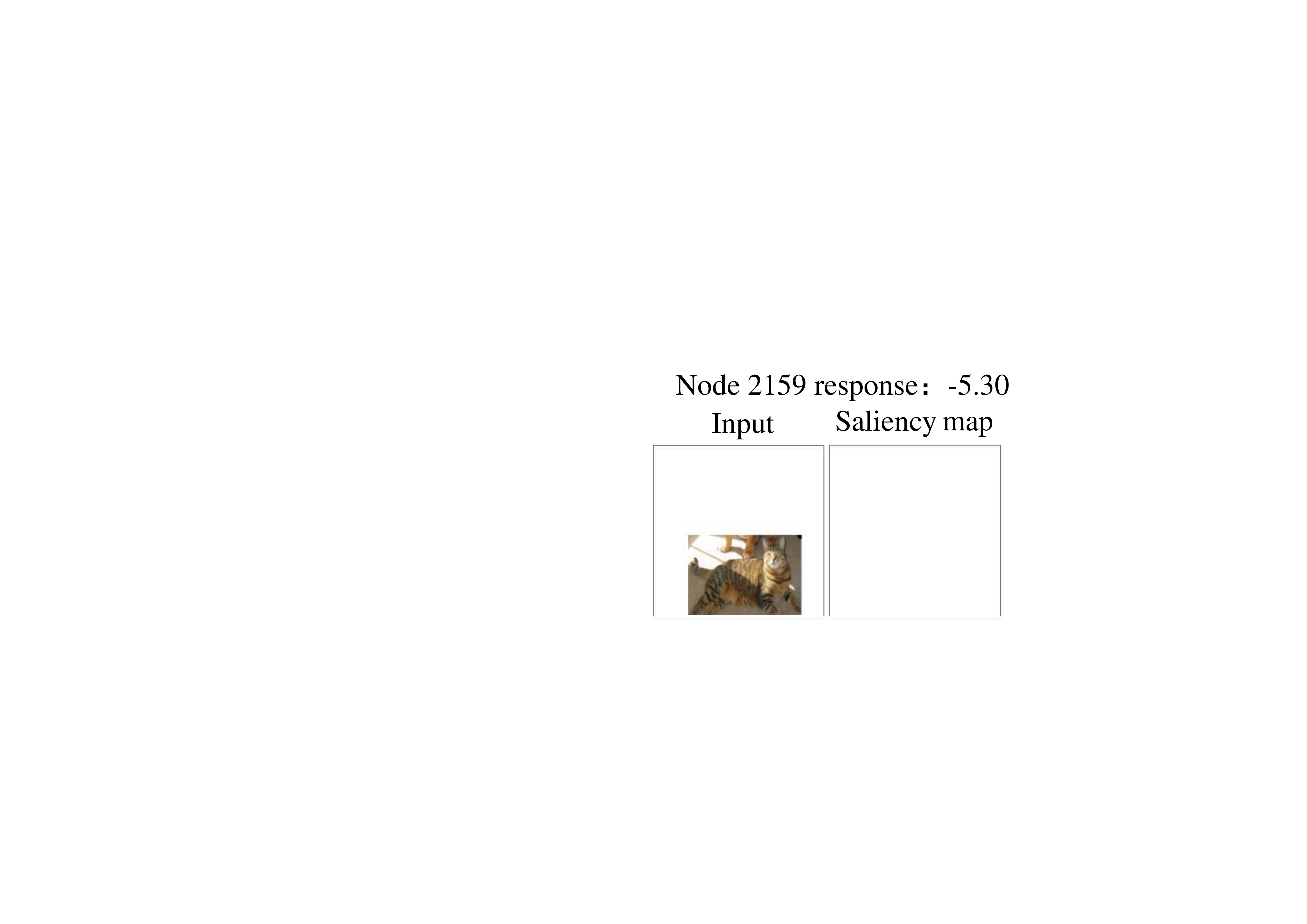}
    }
    \caption{{Analysis of network nodes with entangled semantics. In each test, a feedforward is performed to obtain the node response. Then the saliency is backpropagated from the node to the top Conv.\  layer. Note that in (d), the backprop generates a blank map because of an inactivated state of the node, but we retain the negative response value for a better understanding. }
    }
    \label{fig:coupled node}
\end{figure}

\begin{figure}[t]\centering
\subfigure[Conceptual diagram of node explanation]{
\includegraphics[width=\linewidth]{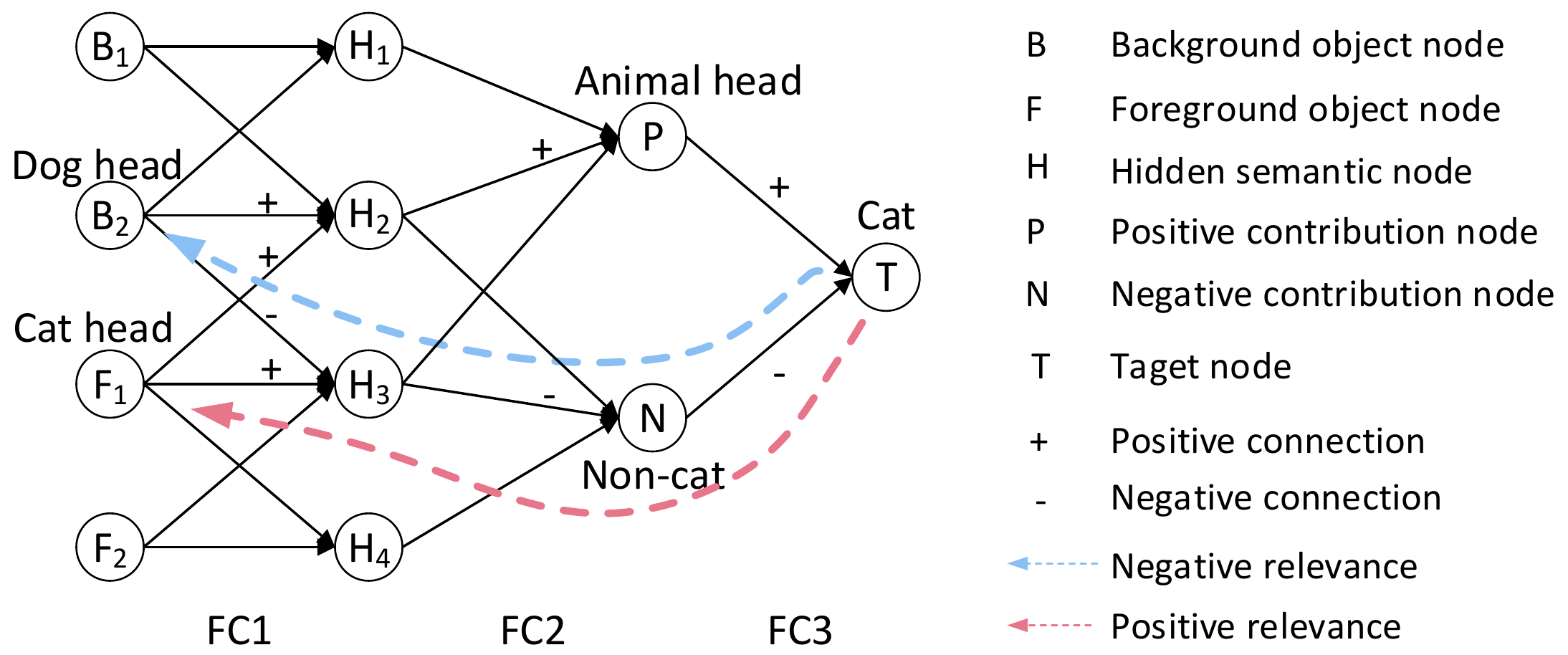}
}
\hspace{0.8cm}
\subfigure[Backprop from negative contribution nodes]{
\includegraphics[width=6.4cm]{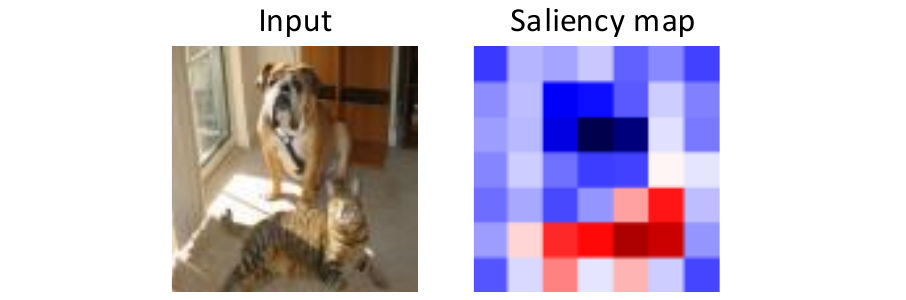}
}
\caption{An example of explanation for network nodes. 
The positive and negative relevance to the target are respectively marked in the red and blue colors\protect\footnotemark[1].
An even number of negative connections make a positive contribution, depicted in the red dotted arrow. 
}
\label{fig:neg nodes}
\end{figure}

\begin{figure}[t]
\centering
\subfigure[Saliency maps of the original gradient backprop]
{\includegraphics[width=7.8cm]{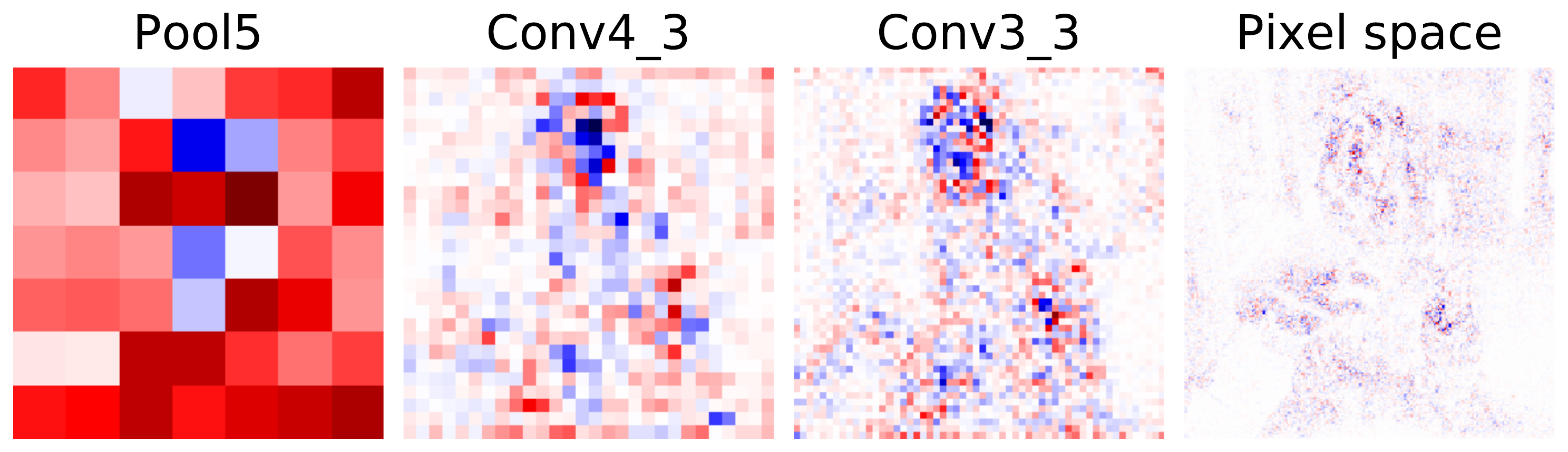}}
\hfil
\subfigure[Saliency maps of TSGB]
{\includegraphics[width=7.8cm]{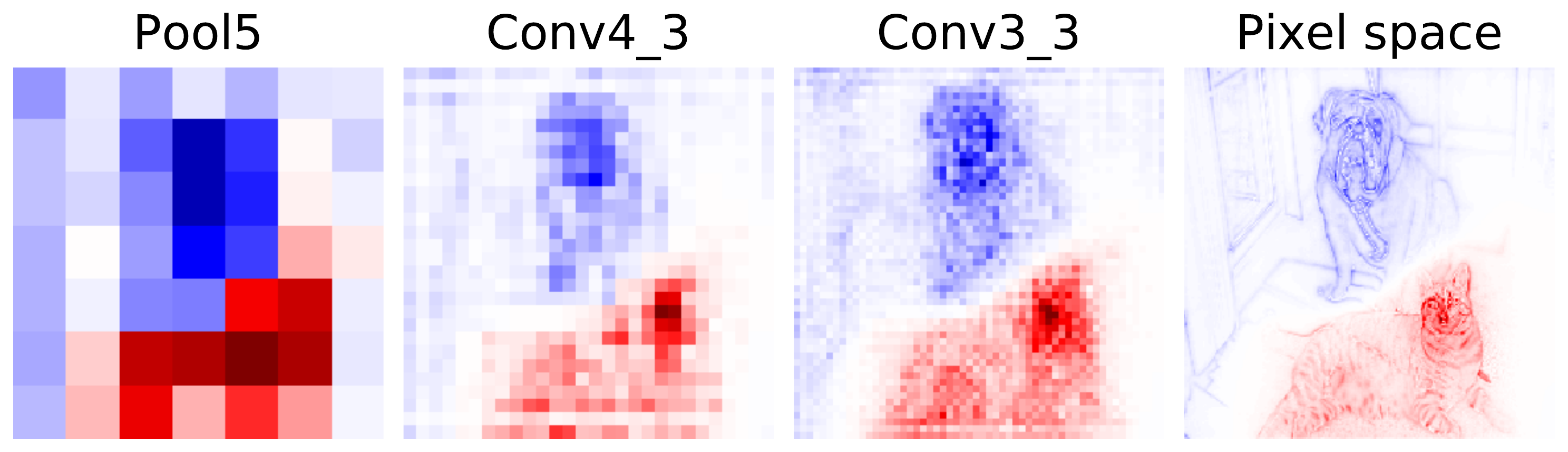}}
\caption{{Comparison of backprops using the original gradient and the proposed TSGB. Both propagations are from the target output node ``tiger cat'' down to low layers. 
We input the same image
in Fig.~\ref{fig:neg nodes}(b) in this test.
}}
\vspace{-2ex}
\label{fig:comparison of grad with TSG}
\end{figure}

\textit{1) Entangled Semantics in the FC Layer:}
In the following, we take the VGG16 model as an example.
As shown in Fig.~\ref{fig:neg nodes}(a), one may intuitively think that the positive contribution nodes (with positive connections) to an output node ``tiger cat'' should encode the ``cat'' related semantic information, e.g., the cat head, the cat tail, etc. However, in practice, when testing on a positive contribution node (Fig.~\ref{fig:coupled node}(a, b)), i.e., the \textit{3258-th node} in the input of the FC3 layer,
both ``dog head'' and ``cat head'' can activate the node.  
Meanwhile, the saliency regions with corresponding semantics are produced by the backprop from the node.
Thus we naturally consider that positive contribution nodes encode entangled semantics, e.g., the ``animal head'', the range of which is even broader than that of the output node's semantics (Fig.~\ref{fig:neg nodes}(a)).
When attributing the target class to the lower layers, passing gradients through these entangled nodes severely affects the target selection, as shown in Fig.~\ref{fig:comparison of grad with TSG}(a) ``Pool5''.

On the other hand, a negative contribution node, i.e., the \textit{2159-th node} in the input of the FC3 layer, is further tested, as illustrated in Fig.~\ref{fig:coupled node}(c, d). We observe that the node's response value is negative for a cat 
being
the input, whereas 
it is
positive for a dog and background 
being
the input. Thus, this node may encode the ``non-cat'' information.
This suggests that the negative contribution is also important to help the network  make a right decision. Furthermore, we surprisingly find that the negative contribution nodes in the FC3 layer contain
the class information, 
as shown in Fig.~\ref{fig:neg nodes}. 
Specifically, using all final negative contribution nodes can result in a class-discriminative saliency map (Fig.~\ref{fig:neg nodes}(b)), which concentrates on the target and suppresses the background. The reasonable explanation is the transformation of gathering the connections with negative signs (i.e., even number of negatives make a positive). For example, the ``cat head'' is negatively relevant to the ``non-cat'' and the ``non-cat'' is negatively relevant to the ``cat'', leading that the ``cat head'' is positively relevant to the ``cat'' (Fig.~\ref{fig:neg nodes}(a)).

\textit{2) Backprop Noise in the Conv.\  Layer:}
As illustrated in Fig.~\ref{fig:comparison of grad with TSG}(a), the gradient backprop generates a lot of noise, losing the target concentration. 
A similar result is also observed in~\cite{smilkov2017smoothgrad,gradcam}.
One possible reason can be explained as follows.
The gradient can be regarded as an approximation to the importance score assigned to per feature. 
Conventionally, model parameters in the Conv.\  layers are trained for the feedforward to extract features. 
Here, in the procedure of the gradient backprop, directly using the original parameters to compute the 
saliency in convolutions (i.e., deconvolution operations) may introduce biases. This is more severe than the situation in the FC layers, because of dozens of local perceptions inside the
convolutions. 
Moreover, the biases are accumulated layer by layer, leading to 
increasing noise along with the gradient backprop,
which prevents achieving a fine-grained explanation.

\footnotetext[1]{This color setting can better distinguish the preserved negative values from positive values in the analysis, which differs from that in the experiment.}

\section{Methodology}\label{sec:method}
\begin{figure*}[!t]\centering
\includegraphics[width=\linewidth,height=0.45\linewidth]{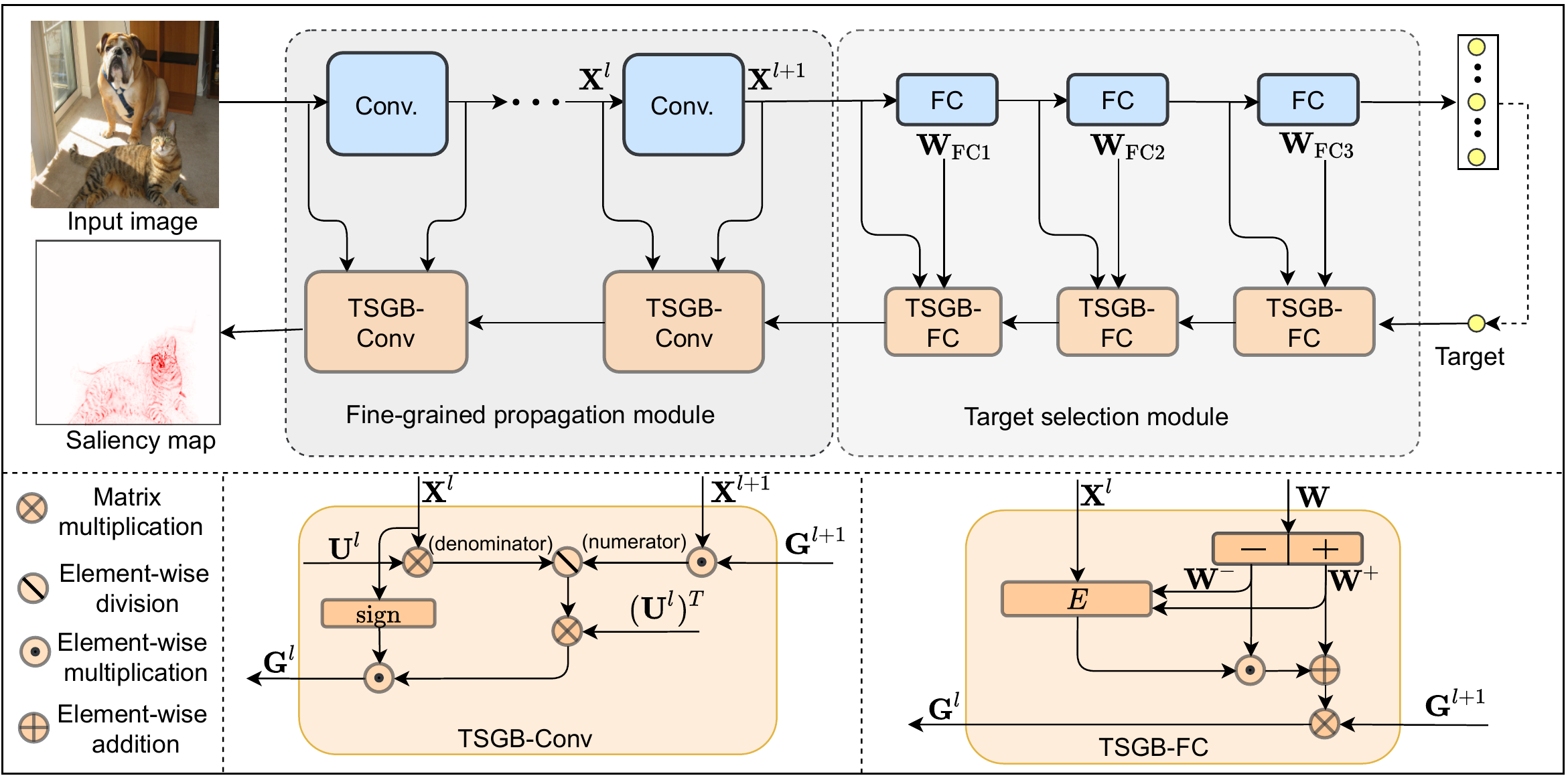}
\caption{{The pipeline of the proposed target-selective gradient backprop (TSGB). Here, we use the VGG network as an example.
}
}
\label{fig:overview}
\end{figure*}

Based on the above analysis, 
we propose a novel CNN visual saliency method, i.e., target-selective gradient backprop (TSGB), 
composed of a target selection module and a fine-grained propagation module, as shown in Fig.~\ref{fig:overview}. 
For a pre-trained CNN model, the FC layers usually encode high-level semantic features related to the target classes, while the Conv.\  layers encode local features related to the object details. Given this prior knowledge, we design two modules of TSGB separately for the FC layers and Conv.\  layers.
We will detail these two modules in the following.

\subsection{Target Selection Module}

According to the analysis of ``entangled semantics'', we propose a target selection module for the FC layers to 
select the target and suppress the target-irrelevant background.

Let $g_i^l$ denote the target-selective gradient (TSG) of the $i$-th node in the $l$-th layer, and $g_j^{l+1}$ denote the propagated gradient of the $j$-th node in the $(l+1)$-th layer. Additionally, the normal gradient $\tilde{g}_i^l = \sum_j {w_{ij} g_j^{l + 1}}$ is given for reference. 
Firstly, given an input image and a pre-trained CNN, we perform a forward propagation and obtain the output scores before the $\mathrm{softmax}$ function. We set the initial gradient of the target node $c$ in the output layer to 1, (i.e., $g_{j=c}^{l+1}=1$), and set the rest nodes' initial gradients to 0, (i.e., $g_{j\neq c}^{l+1}=0$). Then, we compute the TSG layer by layer in a top-down manner.
In the final FC layer, the TSG of the lower layer $g_i^l$ is calculated
by enhancing the negative connection:
\begin{equation}\label{eq:fc}
g_i^l = \sum_j ({w_{ij}^{+}  + E_j(x^l,w)w_{ij}^ - ) g_j^{l + 1}},  
\end{equation}
where $w_{ij}$ is the connection weight, and $w_{ij}^ += \mathrm{ReLU}(w_{ij})$, $w_{ij}^ -=w_{ij}-w_{ij}^ +$. Let $x_i^l$ denote the feature of the $i$-th node in the $l$-th layer. 
The enhancement factor $E_j(x^l,w)$ is 
obtained by the ratio of positive contributions to negative contributions:
\begin{equation}\label{eq:fc_lambda}
E_j(x^l,w)  = \alpha \frac{{\sum_i  {{x_i^l}} w_{ij}^{+} }}{{\sum_i { |{x_i^l}} w_{ij}^- |}}.
\end{equation}
When the positive contribution is larger, or the negative contribution is smaller, the relative entangled strength $\sum{}_i ~ {x_i^l} w_{ij}^ + / \sum{}_i ~ {|x_i^l} w_{ij}^ - |$ will be larger, thereby leading to a larger ratio. 
$\alpha$ is a positive scale coefficient, which adjusts the enhancement ratio. 
It can be deduced that the ratio $\sum{}_i ~ {x_i^l} w_{ic}^ + / \sum{}_i ~ {|x_i^l} w_{ic}^ - |$ for the target $c$ is always larger than $1$ if the output of the target $c$ is positive, 
as
$(\sum{}_i ~ {x_i^l} w_{ic}^ + - \sum{}_i ~ {|x_i^l} w_{ic}^ - |)>0$. 
However, if the ratio is much larger than 1, it may result
in too strong suppression for the foreground objects. Thus, we use the scale coefficient to slightly adjust the enhancement ratio.

\begin{figure}[!t]\centering
\includegraphics[width=\linewidth]{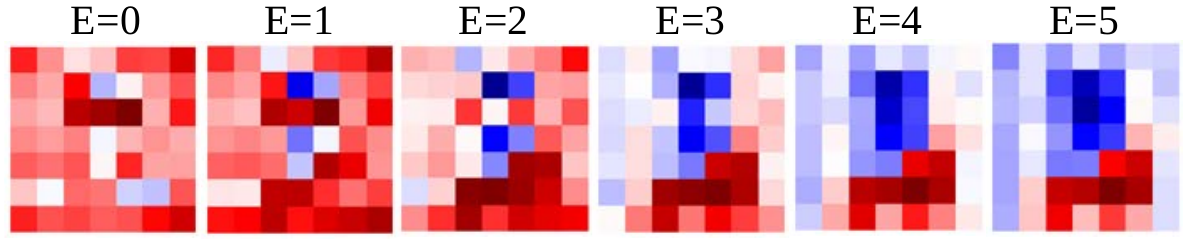}
\vspace{-4ex}
\caption{The influence of the enhancement factor $E$. The saliency maps are from the “Pool5” layer for the target “tiger cat” as in Fig. 5.
}
\label{fig:changeE}
\vspace{-2ex}
\end{figure}

Note that we only rectify the gradients in the final FC layer, and calculate the gradients with the original weights, i.e., $E=1$, for the other FC layers if there exist, such as in the VGG net. This is because that the other FC layers' information is integrated into the final layer's input, which is included in Eq.~\eqref{eq:fc_lambda}.
Different from EBP~\cite{ebp} which only uses positive weights, we make use of both positive and negative weights
for the other FC layers, as both of them
are necessary for the whole module to select the target and suppress the background. 
For example, we use the proposed module to produce the saliency map of ``Pool5'' layer, which is the input layer of FC layer.
As shown in Fig.~\ref{fig:changeE}, we can observe that the target is gradually disentangled from the irrelevant classes and background when the enhancement factor increases.    

\subsection{Fine-Grained Propagation Module}

In this subsection, we further propose a fine-grained propagation module for the Conv.\ layers to efficiently propagate the saliency to the input image space.

In the Conv.\  layers, there exists the local perception over each space location, which is different from the FC layers. Considering this difference, we implement the backprop of Conv.\  layers differently. 
The TSG of the lower layer $g_i^l$ in the Conv.\  layers is devised as
\begin{equation}\label{eq:conv}
g_i^l = \mathrm{sign}(x_i^l)\sum\limits_j {\frac{{x_j^{l + 1}u_{ij}}}{{\sum_i {|x_i^lu_{ij}|} }}} g_j^{l + 1},
\end{equation}
where $u_{ij} = 1$ if $x_i^l$ is inside the receptive field of $x_j^{l+1}$, and $0$ otherwise. 
The denominator is actually the convolution operation with the kernel, each of whose elements is 1. $\mathrm{sign}(\cdot)$ is 
the sign 
function. We leverage the information of feature maps rather than model parameters to propagate the saliency map to the pixel space.
This is because that feature maps are dependent on the input instance, while model parameters are input-agnostic. Feature maps are more accurate for assigning the importance score per feature for a specific instance during propagation. 
As Eq.~\eqref{eq:conv} shows, given an identical input feature, a larger output response indicates the stronger relevance of the input feature to the output feature, leading to a larger TSG.
Note that although no model parameters are explicitly included in the equation, the TSG is related to
model parameters as well. Actually, the computation of the feature $x_j^{l+1}$ is determined by model parameters, which are implicitly contained in Eq. \eqref{eq:conv}.

Eq. \eqref{eq:conv} can be rewritten in a tensor form. Let $\mathbf{X}^l \in \mathbb{R}^{M \times H_l \times W_l}$ and $\mathbf{X}^{l+1} \in \mathbb{R}^{N \times H_{l+1} \times W_{l+1}}$ denote the feature maps in the $l$-th and $(l+1)$-th layer, respectively. $M$ and $N$ are the channel numbers of $\mathbf{X}^l$ and $\mathbf{X}^{l+1}$, respectively. $\mathbf{U}^l \in \mathbb{R}^{M \times N \times K_h \times K_w}$ is a set of defined Conv.\  kernels with the spatial size of $K_h \times K_w$ in the $l$-th layer ($\mathbf{U}^l$ has the same dimension as
the original weight). $\mathbf{G}^l$ and $\mathbf{G}^{l+1}$ are the
TSG maps
in the $l$-th and $(l+1)$-th layers,
respectively. The $m$-th map $\mathbf{G}_m^l 
\in \mathbb{R}^{H_l \times W_l}$ 
is formulated as
\begin{equation}\label{eq:conv2}
\begin{split}
\mathbf{G}_m^l&=
\frac{{\mathbf{X}}^{l + 1} \odot \mathbf{G}^{l + 1}}{{|{\mathbf{X}}^l|\ast {\mathbf{U}^l}}} \ast
(\mathbf{U}_m^l)^T 
\odot~ 
\mathrm{sign}(\mathbf{X}_m^{l}),
\end{split}
\end{equation}
where $\odot$,~$\ast$, and $|\cdot|$ denote the element-wise multiplication, convolution operation, and element-wise absolute value operation, respectively. In our formulation, all elements in $\mathbf{U}$ are ones. 

Let $\mathbf{u} \in \mathbb{R}^{K_h \times K_w}$ denote a single channel of Conv.\ kernel in $\mathbf{U}$. To speed up the computation, we further obtain the following derivation from Eq.~\eqref{eq:conv2}:
\begin{equation}\label{eq:s_conv2}
\begin{split}
\mathbf{G}_m^l 
&=
\big( \sum\limits_{n = 1}^N {\frac{{{{\mathbf{X}_n}^{l + 1}} \odot {{\mathbf{G}_n}^{l + 1}}}}{{|{\mathbf{X}^l}|\ast {\mathbf{U}^l}}}} \big) 
\ast (\mathbf{u}^l)^T
\odot~ 
\mathrm{sign}(\mathbf{X}_m^{l})\\
&=
\frac{{\sum_{n = 1}^N {{\mathbf{X}_n}^{l + 1} \odot {\mathbf{G}_n}^{l + 1}} }}
{{\sum^M_{m = 1} {|\mathbf{X}_m^l|} 
\ast 
\mathbf{u}^l}}
\ast (\mathbf{u}^l)^T
\odot~ 
\mathrm{sign}(\mathbf{X}_m^{l}).\\
\end{split}
\end{equation}
\begin{figure*}[t]\centering
\includegraphics[width=0.95\linewidth]{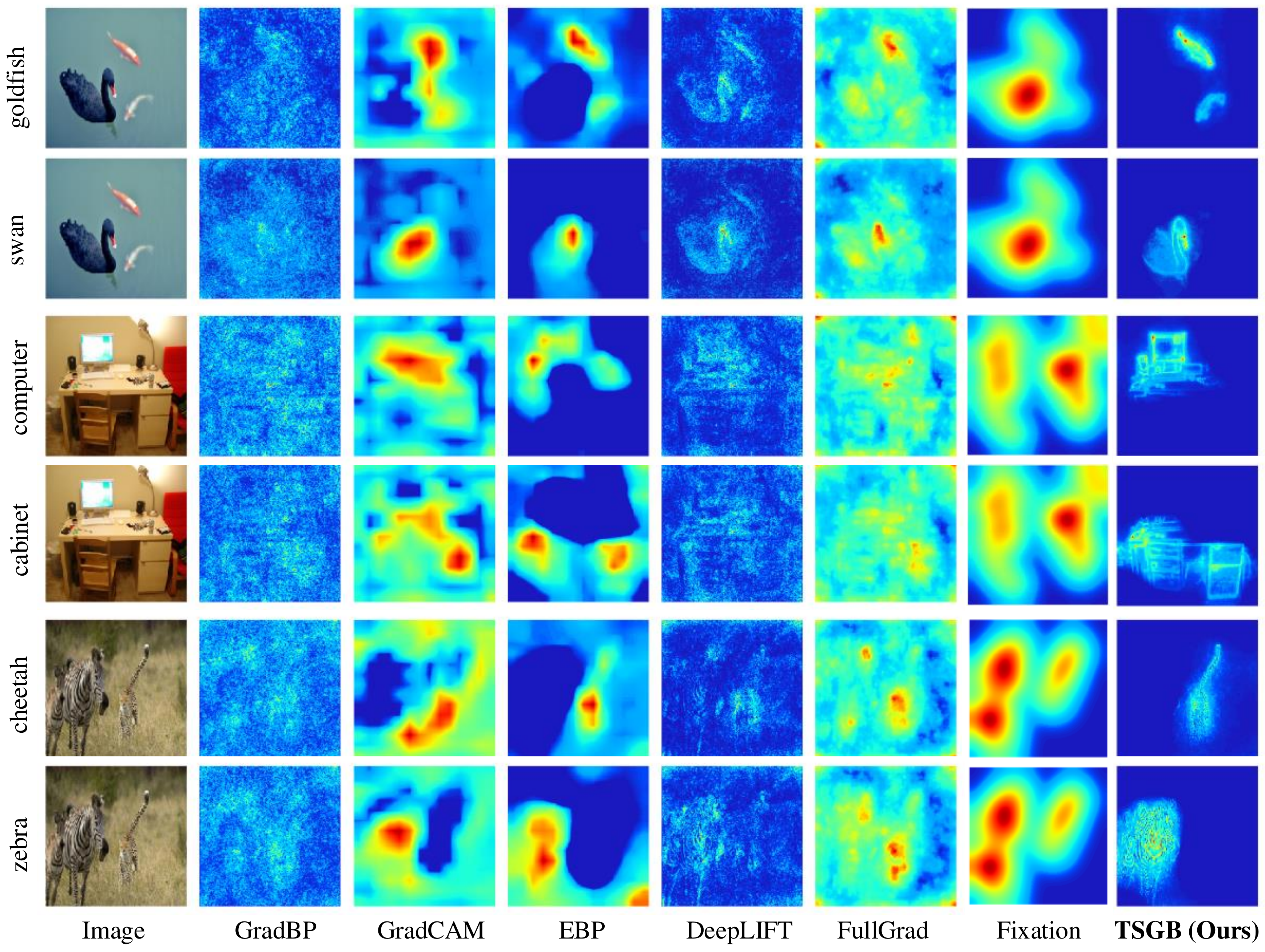}
\caption{{Comparison of different methods on different samples. The saliency maps are generated from different targets, annotated on the left side, in each sample. The deep blue color represents the background, and all other colors represent varying degrees of the target evidence. The negative values are truncated for better contrast.}
}
\vspace{-2ex}
\label{fig:Comparison of Methods}
\end{figure*}
Note that in Eq.~\eqref{eq:conv2}, each channel in ${\mathbf{U}_m^l}$ is equal, leading to obtaining the first line in Eq.~\eqref{eq:s_conv2}. Similarly, considering the first line in Eq.~\eqref{eq:s_conv2}, each channel in ${\mathbf{U}^l}$ is equal leading to each channel in the result of ${|{\mathbf{X}^l}|\ast {\mathbf{U}^l}}$ equal, and thereby obtaining the second line. 
By this transformation, the convolution operation is turned from multiple channels to one channel.
Here, we further analyze the computation complexity of the equation.

\textit{Computation Complexity:} For convenience, we ignore the difference between the multiplication and addition operations, as well as the difference of space scale between input and output layers. The computation complexity of Eq.~\eqref{eq:conv2}, depends on the term $|{\mathbf{X}^l}|\ast {\mathbf{U}^l}$, 
and
thereby 
the computation complexity is
$O(M \times N\times H\times W\times K\times K)$. On the other hand, the computation complexity of the second line in Eq.~\eqref{eq:s_conv2} depends on the term ${\sum^M_{m = 1} {|\mathbf{X}_m^l|}\ast\mathbf{u}^l}$, with the computation complexity being $O(M \times  H\times W\times K\times K)$. Thus, the transformation of Eq.~\eqref{eq:s_conv2} reduces the computation cost $N$ times for 
$\mathbf{G}_m^l$,
and $M\times N$ times for $\mathbf{G}^l$.

\textit{Other Layers:} We formulate the backprop of Normalization layer, including the Batch Normalization layer and the Local Response Normalization layer, as
\begin{equation}\label{eq:other layer}
g_i^l = \frac{{x_j^{l + 1}}}{{x_i^l}}g_j^{l + 1}.
\end{equation}
Eq. \eqref{eq:other layer} is also utilized for the backprop of a type of Average Pooling layer whose input features contain negative values, such as in the case of DenseNet. Otherwise, we directly use the original gradient operations for the other layers in CNNs, including ReLU, Max Pooling, Adaptive Pooling, Skip Connection, Concat layer, and common Average Pooling layer, etc.

As shown in Fig.~\ref{fig:comparison of grad with TSG}(b), this fine-grained propagation module can effectively deliver the TSG to the image space to generate high-resolution saliency maps,
meanwhile keeping the target concentration. 
Note that the TSG can be propagated to any layer inside the network to analyze the attributions of channels of interest according to different demands of semantic levels and spatial scales.

\section{Experiments}\label{sec:expt}
In this section, we first qualitatively validate the proposed TSGB via visual comparisons. Then in quantitative experiments, we evaluate the proposed TSGB with weakly-supervised localization tasks on the ImageNet dataset~\cite{Russakovsky2015ImageNetLS} and the Pascal VOC dataset~\cite{VOC2014ThePV}.
Furthermore, we evaluate the faithfulness of the explanations with pixel perturbation~\cite{RISE_petsiuk2018rise} and sanity check~\cite{Sanity}.
Finally, we perform the bias diagnosis, 
the medical image test
and the ablation study.

We compare our method with
several other competitors including
GradBP~\cite{GradBP}, GradCAM~\cite{gradcam}, DeepLIFT~\cite{LIFT_shrikumar2017learning,18ICLR}, EBP~\cite{ebp}, FullGrad~\cite{fullgrad} and Fixation~\cite{TIP_fixation}.
These competitors are the state-of-the-art saliency methods in the single backward pass type, which is consistent with the type of our method.
For our method, we set the scale coefficient to 0.5$\sim$1.3 for the negative enhancement in Eq.~\eqref{eq:fc_lambda}. More details can be found in Section \ref{Ablation}. 
We follow the processing in~\cite{fullgrad} to obtain final saliency maps by first multiplying the produced target-selective gradients to feature maps, and then summing all the elements along the channel dimension.

\subsection{Visual Comparison}
\textit{1) Comparison on Different Samples:}
We employ TSGB to generate saliency maps from different targets on different samples in comparison with the other competitors.
As shown in Fig.~\ref{fig:Comparison of Methods}, GradBP and DeepLIFT generate noisy maps, which highlight most foreground objects, even including some target-irrelevant objects.
FullGrad focuses on the most dominant objects rather than the target. Fixation only generates almost the same saliency maps w.r.t. different targets of each image.
One reasonable explanation for Fixation is that the backprop in the FC layers neglects the negative connections, leading to the lack of the target-selectiveness.
GradCAM and EBP can produce  class-discriminative maps.
However, their results still contain irrelevant backgrounds, especially on the borders of images, such as in the cases of ``goldfish'', ``cabinet'', ``cheetah'' and ``zebra''. It is also worth mentioning that the generated saliency maps from FullGrad, GradCAM, and EBP are coarse. In contrast, TSGB can produce target selective and fine-grained maps with clear targets' 
boundaries
and fewer irrelevant backgrounds. Furthermore, the explanatory results of TSGB are more 
human 
interpretable, when compared to its competitors.

\begin{figure*}[!t]\centering
\includegraphics[width=0.86\linewidth]{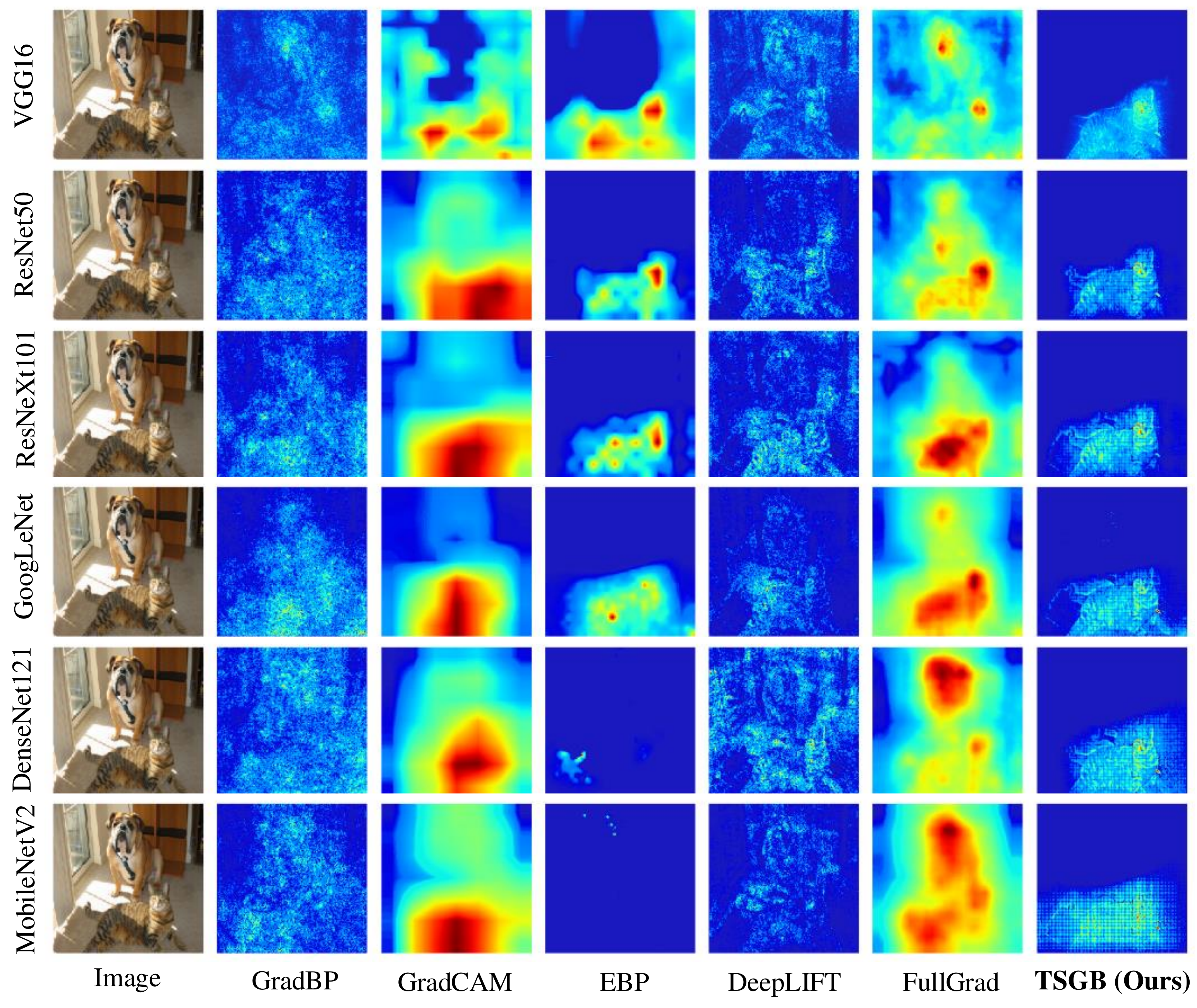}
\vspace{-2ex}
\caption{{Comparison of different methods on different models. The models' names are annotated on the left side. The saliency maps are generated from the same target, i.e., ``tiger cat''.}
}
\vspace{-2ex}
\label{fig:Comparison of Models}
\end{figure*}

\textit{2) Comparison of Different Models:}
To verify the generalization of the proposed TSGB, we further conduct the experiments across various CNN models, along with the competitors. From Fig.~\ref{fig:Comparison of Models}, we can find that over several cases EBP can produce saliency maps with fewer background areas than GradBP, DeepLIFT and FullGrad, while 
it
totally fails on DenseNet121 and MobileNetV2.
The main reason is that the features in DenseNet121 and MobileNetV2 contain negative values, which affects the robustness of EBP.
GradCAM is valid for these 
tested
models, while its results cannot discriminate the borders of targets precisely. 
Moreover, GradCAM also fails 
like
EBP if the gradients are backpropagated to the low layer~\cite{gradcam}. Unlike these competitors, TSGB 
shows its advantage of being
target-selective, fine-grained, and robust for extensive models, even for the models containing negative-value features (i.e., DenseNet121, MobileNet, etc.).
In addition, we  find that saliency maps generated on ResNet50 and VGG16 are better than the other
models.
The target saliency on MobileNetV2 is relatively blurry when compared to the other networks. 
One possible reason is that the computation-efficient model cannot learn good features as discriminative as other conventional models.
Since the official code of Fixation does not support the models of ResNet50, ResNeXt, DenseNet and MobileNet, we omit the evaluation of Fixation on these models.

\subsection{Weakly-Supervised Localization}\label{subsec:weak loc}

\begin{table}[t]
\caption{{Localization on the ImageNet val set (lower is better). Error rates of GradBP, GradCAM, and EBP for VGG16 are taken from~\cite{gradcam}. 
DeepLIFT refers to the ``Captum'' package in Pytorch1.4~\cite{Pytorch}. Fixation refers to the official code in~\cite{TIP_fixation}.
}}
\label{table:imagenet}
\begin{center}
\tabcolsep0.03in
\begin{tabular}{c c c c c}
\toprule
       &VGG16                &     & ResNet50           & \\
\cmidrule(lr){2-3} \cmidrule(lr){4-5}
Method & Top5 LOC error (\%) & FPS & Top5 LOC error (\%) & FPS\\
\midrule
GradBP~\cite{GradBP}&51.46&25.64&55.44&\textbf{34.19}\\
GradCAM~\cite{gradcam}&46.41&32.26&40.73&29.63\\
DeepLIFT~\cite{LIFT_shrikumar2017learning}&55.32&7.12&53.11&17.78\\
EBP~\cite{ebp}&63.04&23.26&44.44&26.67\\
FullGrad~\cite{fullgrad}&47.82&12.05&46.35&9.93\\
Fixation~\cite{TIP_fixation}&58.38&0.39&-&-\\
TSGB (Ours)&\textbf{43.46}&\textbf{43.48}&\textbf{40.49}&31.25\\
\bottomrule[1pt]
\end{tabular}
\vspace{-2ex}
\end{center}
\end{table}

\textit{1) Object Localization:} A satisfactory saliency method is expected to generate target-relevant saliency maps, where the areas with high intensity indicate the positions of targets.
Following~\cite{gradcam,ebp,pami/CaoHYWWT19}, we evaluate the visual saliency methods with the
weakly-supervised object localization task on the ImageNet dataset 
using the VGG16 and ResNet50 models, which are pre-trained with the classification labels. 

On the ImageNet 2012 validation (val) set, we first predict categories, and then use saliency methods to generate the saliency maps. The top-5 localization (LOC) error is evaluated under the protocol of the ILSVRC challenge~\cite{Russakovsky2015ImageNetLS}. 
 
After achieving the saliency maps, we search the best performing thresholds for different methods and binarize the saliency maps with the selected thresholds
to obtain the bounding boxes. Besides the binarization, 
we do not append any other post-processing techniques to our method. 
As shown in Table~\ref{table:imagenet}, TSGB outperforms the other methods in localization errors
both
explained models.
For example, the results of TSGB are
43.46/40.49, as compared to 46.41/40.73 of the second-best method, i.e. GradCAM, on VGG16/ResNet50.
Note that GradCAM additionally applies the post-processing technique, i.e., searching for the largest connected component after binarization.
Compared to ResNet, the VGG model has more FC layers, which are possible to involve the stronger entanglement as stated in Section~\ref{sec:Analysis}. In this situation, using TSGB to disentangle the semantics in the FC layers will boost the performance of VGG. 
TSGB outperforms
FullGrad by 4.36\%/5.86\% on VGG16/ResNet50.
FullGrad aggregates all Conv-layer saliency maps to improve the performance but 
it consumes much more computation memory.
We find that most methods achieve lower error rates on ResNet50 than VGG16. This is likely owing to the higher classification capacity of ResNet50, leading to better localization performances.

Moreover, we test the average running speed
on a GeForce GTX 1080 Ti GPU. The proposed TSGB achieves the highest speed at 43 frames per second (FPS), which is 6 times faster than the DeepLIFT (7 FPS). Note that Fixation does not support GPU computation in its backprop, resulting in 
the slow running speed.

\textit{2) Point Localization:} Considering that the explanatory results intend to focus on the most discriminative regions of targets, we use
another popular evaluation metric, Pointing Game~\cite{ebp}, to measure the explanatory results.
This metric is defined as the ratio of hits, where a hit is counted if the maximum point of the saliency map is inside the target region.
As shown in Table~\ref{table:point game}, our method achieves the superior performance over the other methods on the Pascal VOC2007 test set, which can be attributed to the target-selectiveness of TSGB. 
GradBP and Fixation achieve much lower accuracy. This is probably because that the point localization of GradBP is easily interfered by the noise, and Fixation cannot focus on the target class, which is consistent with the visual comparison experiments (see Fig. \ref{fig:Comparison of Methods}).

\begin{table}[t]
\caption{{Pointing Game on the VOC2007 test set (higher is better). The results of GradBP, GradCAM, and EBP are taken from~\cite{ebp}.}}
\label{table:point game}
\begin{center}
\tabcolsep0.03in
\begin{tabular}{c c c c c}
\toprule
       &VGG16                &     & ResNet50           & \\
\cmidrule(lr){2-3} \cmidrule(lr){4-5}
Method & Mean accuracy (\%) & FPS & Mean accuracy (\%) & FPS\\
\hline
GradBP~\cite{GradBP}&76.00&18.18&65.80&\textbf{16.95}\\
GradCAM~\cite{gradcam}&86.60&\textbf{18.52}&90.60&16.25\\
DeepLIFT~\cite{LIFT_shrikumar2017learning}&79.05&7.69&82.72&3.12\\
EBP~\cite{ebp}&80.00&10.41&89.20&6.31\\
FullGrad~\cite{fullgrad}&84.16&8.56&88.99&3.14\\
Fixation~\cite{TIP_fixation}& 74.52& 0.44& -& -\\
TSGB (Ours)&\textbf{89.33}&18.18&\textbf{90.68}&11.82\\
\bottomrule[1pt]
\end{tabular}
\vspace{-2ex}
\end{center}
\end{table}

\subsection{Faithfulness Check}\label{faithful}
\textit{1) Pixel Perturbation:} 
In order to evaluate the faithfulness of explanatory results at pixel level,
we use the deletion metric~\cite{RISE_petsiuk2018rise} to test TSGB. The intuition behind this metric is that if the saliency region is responsible for the model prediction, the prediction probability will descend when erasing the corresponding region.
This protocol is to measure the decline in prediction probability of classification when iteratively perturbing the important pixels according to the rank of saliency values generated by a saliency method.
The steeper the decline
(i.e., the lower deletion score) is, the more reliable the saliency method is.
As shown in Table~\ref{table:perturb}, TSGB achieves the lowest score, which suggests TSGB is the most 
faithful to the model predictions and capable of capturing the fine-grained details corresponding to the targets.

\begin{table}[t]
\caption{{Pixel perturbation on the VOC2012 val set (lower is better). Lower deletion score 
means
higher faithfulness of saliency methods.}}
\vspace{-2ex}
\label{table:perturb}
\begin{center}
\tabcolsep0.1in
\begin{tabular}{c c c}
\toprule
Method & Deletion score & FPS\\
\hline
GradBP~\cite{GradBP}&0.1932&10.10\\
GradCAM~\cite{gradcam}&0.1680&10.10\\
DeepLIFT~\cite{LIFT_shrikumar2017learning}&0.1621&1.42\\
EBP~\cite{ebp}&0.5028&5.41\\
FullGrad~\cite{fullgrad}&0.1667&4.78\\
TSGB (Ours)&\textbf{0.1564}&\textbf{14.29}\\
\bottomrule[1pt]
\end{tabular}
\end{center}
\end{table}

\begin{figure}[t]\centering
\includegraphics[width=0.95\linewidth]{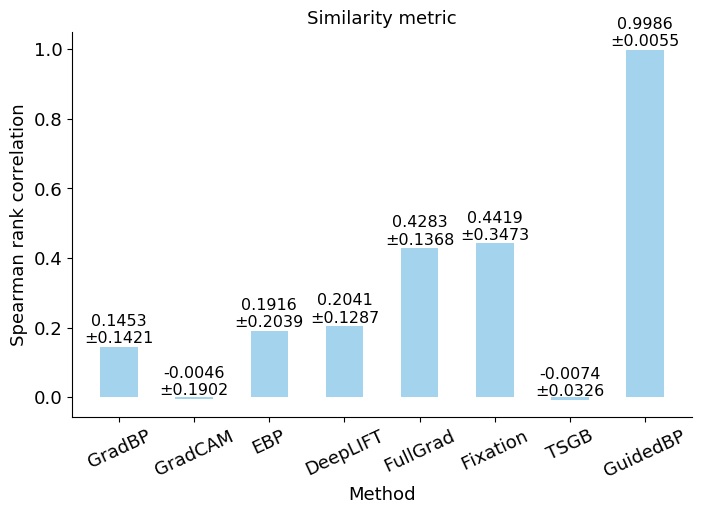}
\caption{{Sanity check with similarity metric for model randomization. Spearman rank correlation is taken as the similarity metric. The values above the bar are the means and standard deviations of similarities between the original explanations and the randomized explanations on ImageNet. Lower similarity denotes better faithfulness of explanations.}}
\vspace{-2ex}
\label{fig:sanity check}
\end{figure}

\textit{2) Sanity Check:} As suggested by~\cite{Sanity}, we conduct the sanity check for the proposed TSGB to validate whether the explanatory results are sensitive to the model parameters or not. If the explanatory results are 
similar before and after the model parameters are randomized, 
the corresponding saliency method is more risky
in trustworthiness.
We evaluate the similarity with Spearman rank correlation before and after the randomization of model parameters for our TSGB and the other comparative methods, including GuidedBP~\cite{GGBP_springenberg2014striving} for reference. As illustrated in Fig. \ref{fig:sanity check}, TSGB and GradCAM are sensitive to the change of the 
parameter values
while GuidedBP is much independent 
of model parameters.

\subsection{Diagnosing Bias and Failure Cases}\label{Diagnosing}

\begin{figure}[t]
\centering
\includegraphics[width=0.9\linewidth]{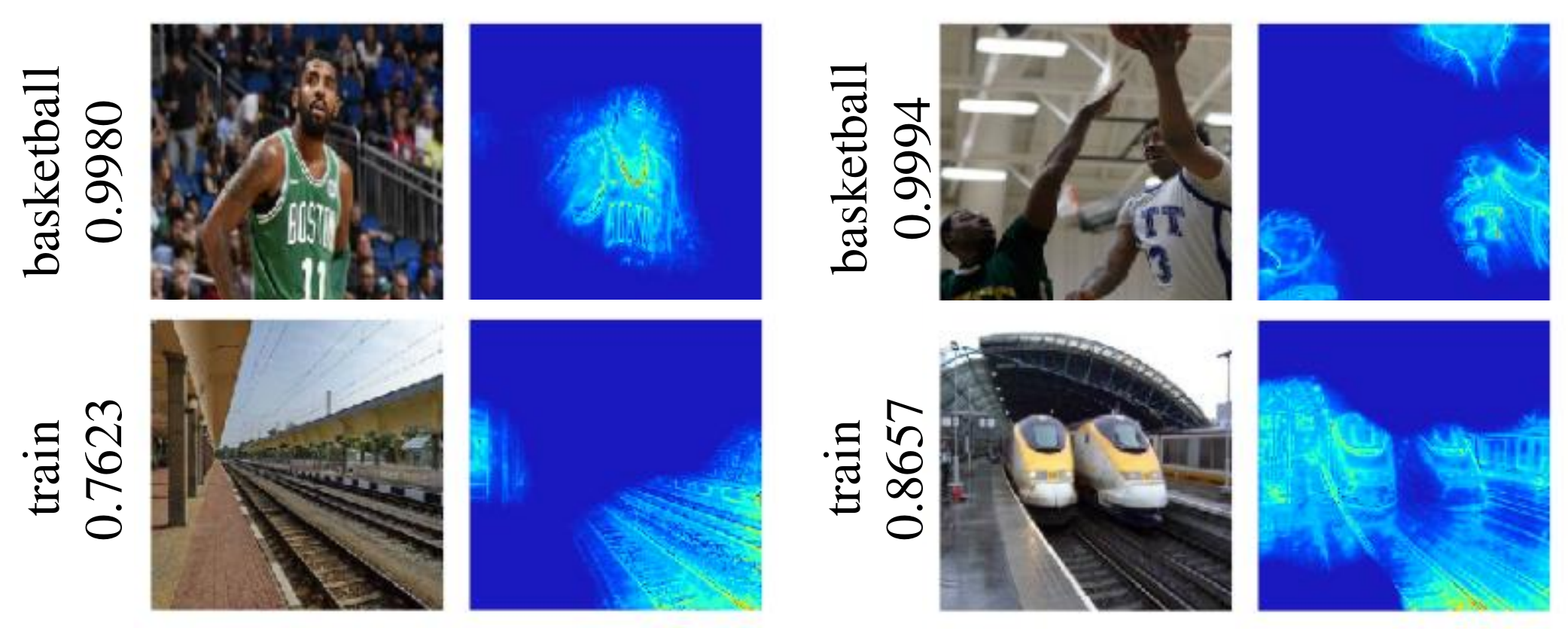}
\caption{{Diagnosing bias and failure cases. The texts denote the predicted targets and possibilities. TSGB can help diagnose the biases in the model and dataset without suppressing the useful information, even in the background.}}
\label{fig:diagnose}
\end{figure}

We adopt TSGB to diagnose the biases in the VGG16 network pre-trained on ImageNet. 
As shown in Fig. \ref{fig:diagnose}, the man is recognized as ``basketball'' (top left), and the station is recognized as ``train'' (bottom left), which makes it difficult to catch
the failure clues by only knowing the prediction possibilities. 
Fortunately, with the help of target-selective and fine-grained saliency maps generated by TSGB, one can easily
understand the reason why the model makes such decisions. For example, the ``basketball'' class is predicted by seeing the sports suit, and the ``train'' class is predicted by seeing the rail.
One reasonable explanation of model biases is that co-occurring objects, e.g., sports suit and basketball, rail and train, exist in the training dataset. 
For the right two cases in Fig. \ref{fig:diagnose}, our method fails to produce the target-specific visualized maps, where some relevant backgrounds are not suppressed. This is because that these backgrounds are involved in the model predictions of the target classes. This also suggests that TSGB is faithful to the model. 

\subsection{Explanation for Medical Images}\label{sec:medical images}

\begin{figure*}[!t]\centering
\includegraphics[width=0.9\linewidth]{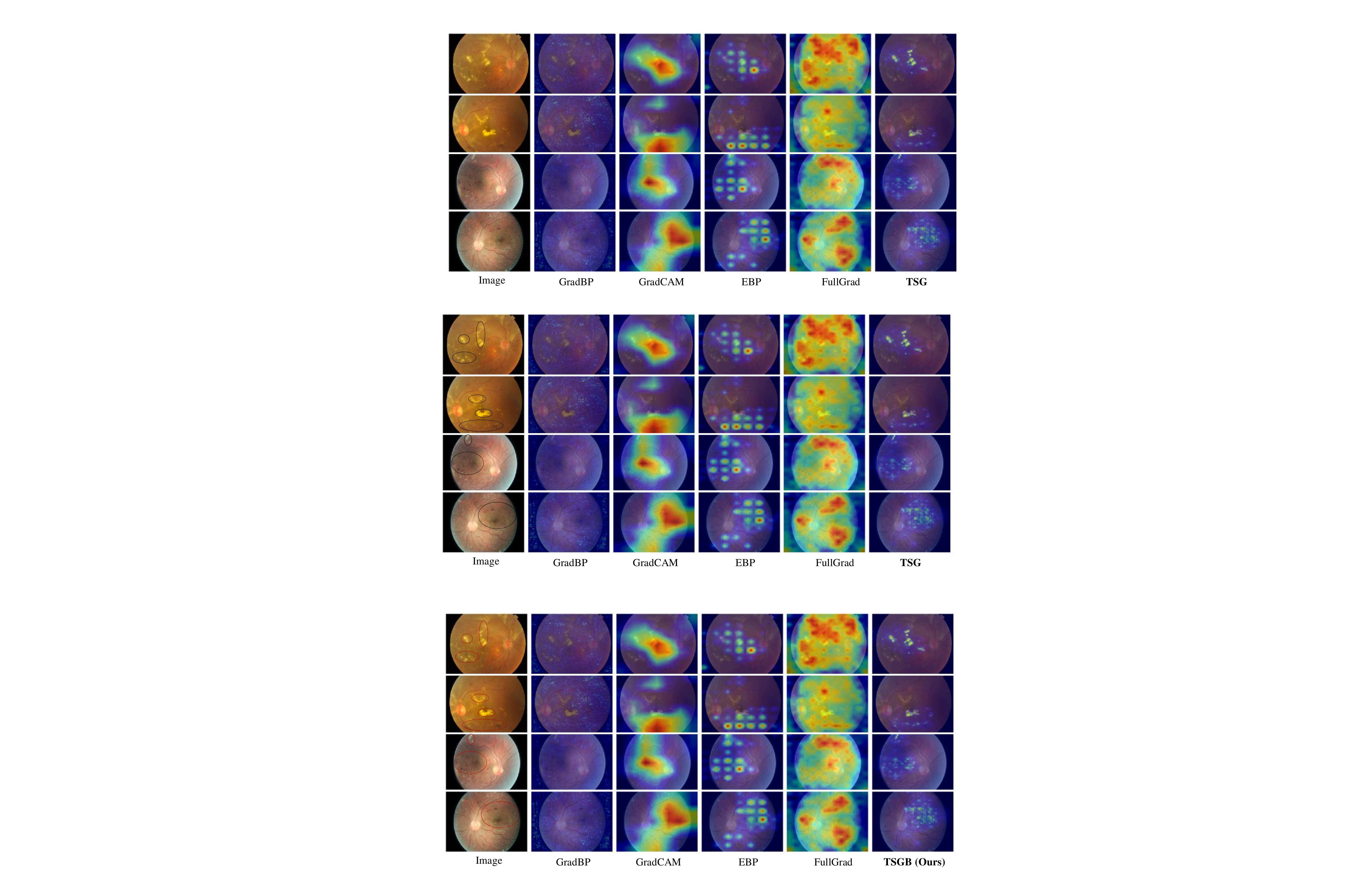}
\vspace{-2ex}
\caption{{Explanation for 
several
medical images. We compare the proposed TSGB with GradBP~\cite{GradBP}, GradCAM~\cite{gradcam}, EBP~\cite{ebp}, and FullGrad~\cite{fullgrad}. The red ovals denote the lesions of retinas.}
}
\vspace{-1ex}
\label{fig:medical images}
\end{figure*}

To test the generalization of TSGB on the different types of images, we use TSGB to explain the deep neural model trained on the medical image dataset, i.e., the Kaggle Diabetic Retinopathy dataset. 
The images in this dataset contain various texture features and color features, which are non-object-like features.
Thus there is a big domain gap between the Kaggle Diabetic Retinopathy dataset and the ImageNet dataset.
The explained model is ResNet152 trained on the Kaggle Diabetic Retinopathy dataset with image-level labels.
It took around more than 100 epochs to train this model to 
achieve
97\% accuracy for classification.
As shown in Fig.~\ref{fig:medical images}, TSGB obtains more reliable explanatory results than the other 
competing
methods. 
Benefiting
from the target-selectiveness, TSGB can focus on the disease-relevant regions. More importantly, with the property of fine-grainedness, TSGB can effectively highlight the detailed patterns in the medical images.

\subsection{Ablation Study}\label{Ablation}
\textit{1) Target Selection Module vs. Fine-Grained Propagation Module:}
We compare the proposed target selection module with the proposed fine-grained propagation module via ablation study on the ImageNet localization task. We choose the vanilla gradient backprop as our baseline. As shown in Table~\ref{table:ablation}, when replacing the vanilla gradient 
backprop 
in the FC layers with the target selection module, the LOC error
achieved by TSGB-FC+Grad becomes
4.39\% lower. When replacing the vanilla gradient 
backprop
in the Conv.\  layers with the fine-grained propagation module, the LOC error
achieved by Grad+TSGB-Conv becomes
1.29\% lower. Although the fine-grained propagation module seems
less useful, when we further append the fine-grained propagation module to the target selection module, the LOC error continues to decrease, i.e., 5.14\% lower than
that obtained by
TSGB-FC+Grad. This shows that both of the proposed modules are necessary to TSGB and provide complementary benefits to TSGB. Those two modules are tied tightly in one framework, achieving better
performance than 
only using
a single module.

\begin{table}
\newcommand{\tabincell}[2]{\begin{tabular}{@{}#1@{}}#2\end{tabular}}
 \centering
   \caption{Ablation study for TSGB (lower is better). ``Grad'' denotes the vanilla gradient
   backprop.
   ``TSGB-FC'' denotes the target selection module.``TSGB-Conv'' denotes the fine-grained propagation module.}
 \label{table:ablation}
 \footnotesize
 \label{Table2}
 \begin{tabular}{lccc}
  \toprule
  {Methods} &~TSGB-FC~ &~TSGB-Conv~ &\tabincell{c}{LOC error (\%)}\\ 
  \hline
  Grad  & & &52.99 \\
  TSGB-FC+Grad & \checkmark& &48.60 \\
  Grad+TSGB-Conv & &\checkmark &51.70 \\
  TSGB (Ours)&\checkmark &\checkmark &\textbf{43.46}\\
  \bottomrule
 \end{tabular}
\vspace{-2ex}
\end{table}

\begin{figure}[t]
\centering
\includegraphics[width=0.8\linewidth]{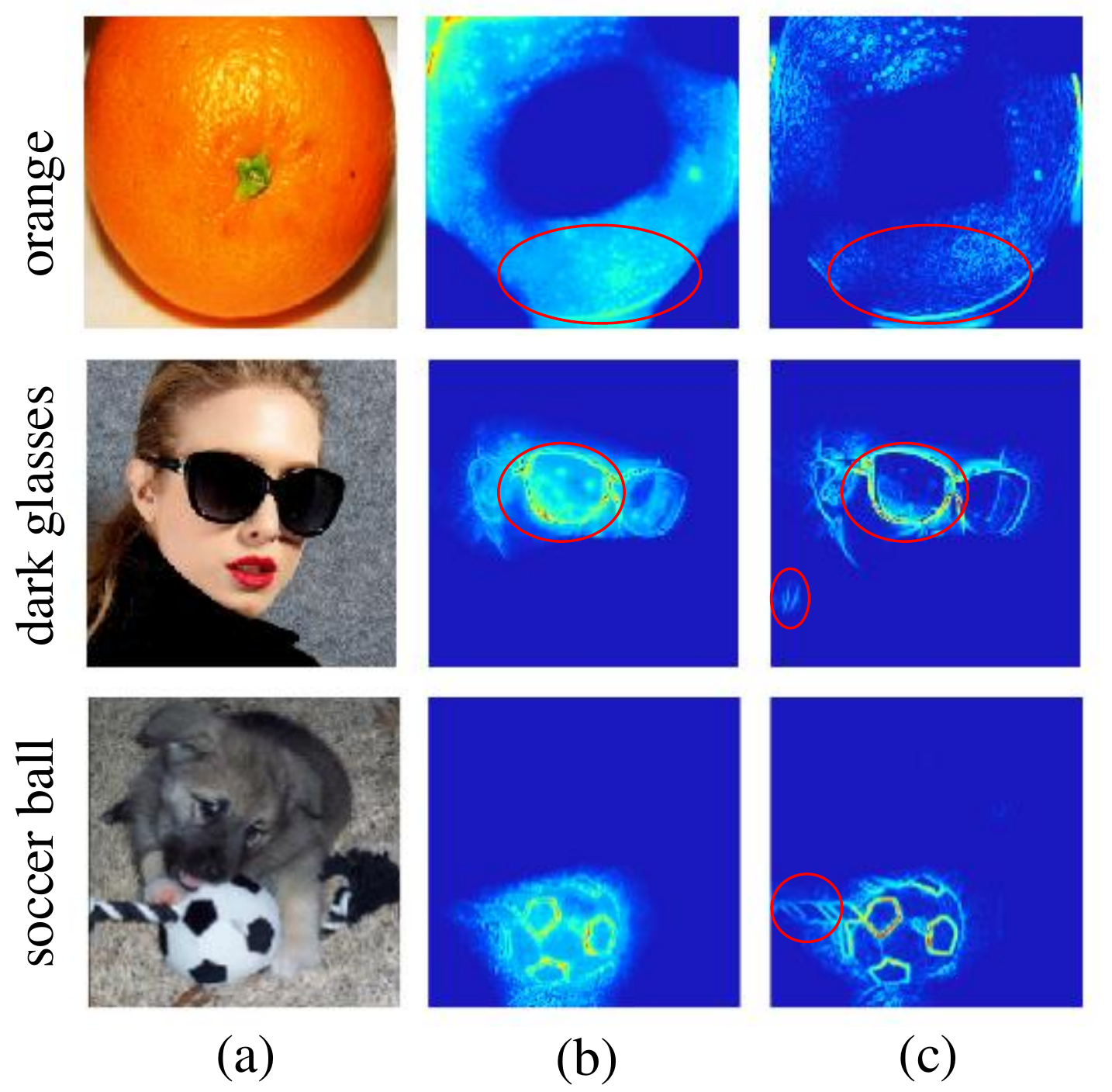}
\vspace{-2ex}
\caption{{
Influence of the fine-grained propagation module and the edge detector.
(a) Input images. (b) Saliency maps generated by TSGB. 
(c) Saliency maps generated by replacing
of the fine-grained propagation module in TSGB with the edge detector. 
}}
\vspace{-2ex}
\label{fig:vs. edges}
\end{figure}

\textit{2) Fine-Grained Propagation Module vs. Edge Detector:} 
Benefiting from the fine-grained propagation module, the saliency maps can highlight clear details relevant to the targets, such as the examples in Fig.~\ref{fig:Comparison of Methods}. When we propagate the saliency maps to the pixel level, the visualizations appear more edges-like patterns, since the low layers in CNN are inclined to extract edge features and other low-level features from the images. Nevertheless, the fine-grained propagation module is not equivalent to an edge detector. We replace the fine-grained propagation module with the edge detector and evaluate the new setting with the pixel perturbation experiment, which turns out 11.06$\%$ worse. Compared with the edge detector, the fine-grained propagation module can also highlight 
the texture and color patterns, besides the edge patterns, such as the ``orange'' and the ``dark glasses'' in Fig. \ref{fig:vs. edges}. 
Moreover, the fine-grained propagation module can refine the details in coarse saliency maps in a top-down manner, such that it further suppresses the irrelevant object parts, such as the hair tail in the ``dark glasses'' and the bar in the ``soccer ball'' in Fig. \ref{fig:vs. edges}. 
In addition, the fine-grained propagation module can propagate the saliency maps to different semantic levels at different spatial scales, and 
it
can analyze the attributions of different features channels. 

\begin{figure}[t]
\centering
\includegraphics[width=0.9\linewidth]{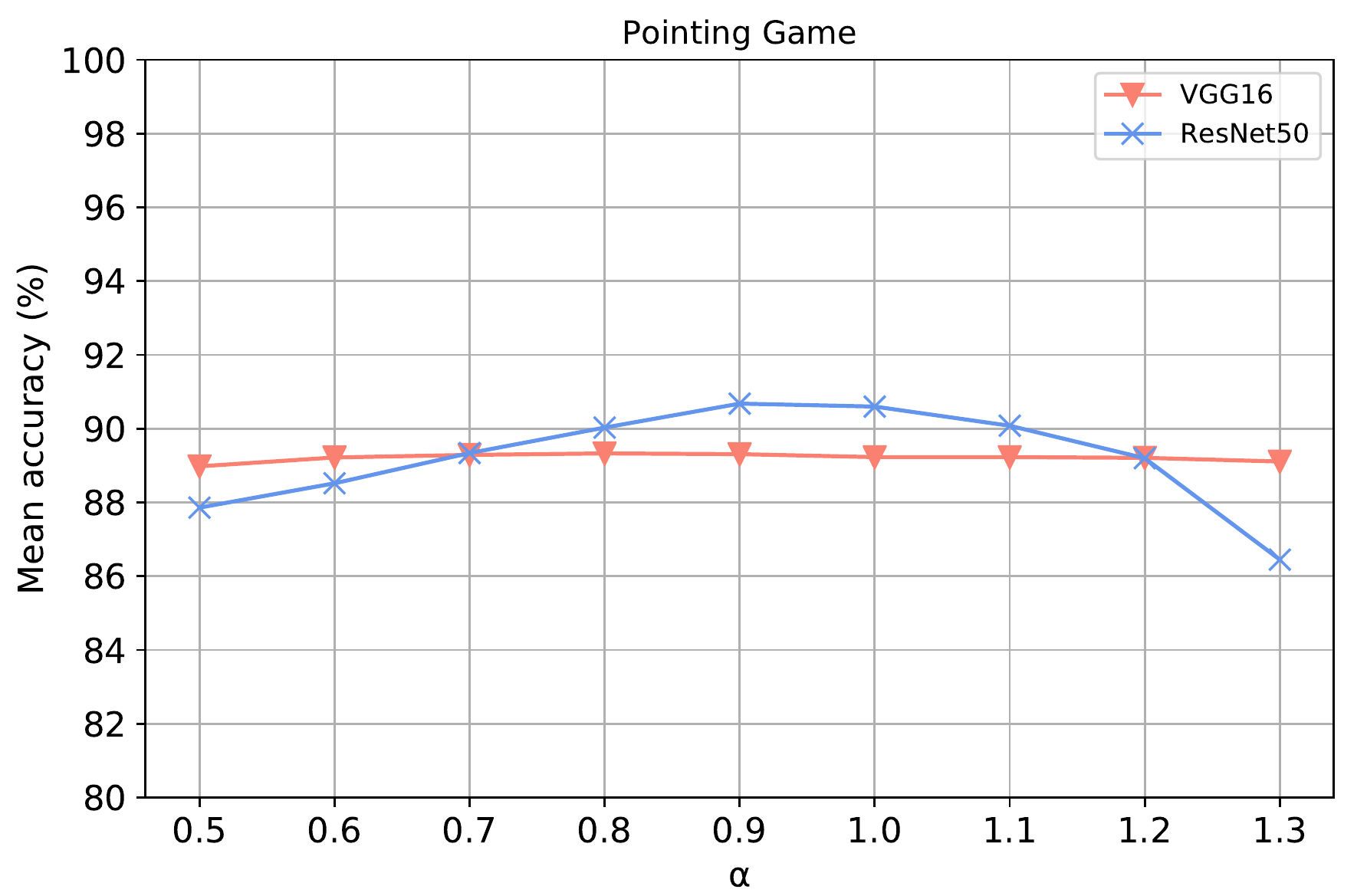}
\vspace{-1ex}
\caption{{Influence of 
different values of the scale coefficient 
$\alpha$ on the performance of TSGB.
}}
\vspace{-1ex}
\label{fig:Parameters analysis}
\end{figure}

\textit{3) Influence of Scale Coefficient:}
To analyze the influence of choosing 
different values of the scale coefficient 
$\alpha$ in Eq.~\eqref{eq:fc_lambda}, we test the proposed TSGB with Pointing Game, as mentioned in ``Point localization'' in Section \ref{subsec:weak loc}. We record the experimental results corresponding to 
varied
$\alpha \in [0.5:0.1:1.3]$ both on the VGG16 and ResNet50 models. As shown
in Fig.~\ref{fig:Parameters analysis}, we can find that each model has one peak mean accuracy as $\alpha$ varies
from 0.5 to 1.3,
where VGG16 obtains the best result at $\alpha=0.8$ and ResNet50 obtains the best result at $\alpha=0.9$.    
Furthermore, the results on VGG16 are less sensitive to the scale coefficient $\alpha$ than those on ResNet50. Specially, the accuracy on the VGG16 model fluctuates within a very small 
extent, i.e., 0.35, in the whole range of $\alpha$. 
On both models, there is less
fluctuation in the mean accuracy for $\alpha \in [0.6,1.2]$. 

\section{Conclusion and discussion}\label{sec:conclusion}
To probe the CNN visual saliency, we propose a novel saliency backprop method,
i.e., target-selective gradient backprop (TSGB), which consists of a target selection module and a fine-grained propagation module.
The target selection module adaptively enhances the negative connections to disentangle the target class from the irrelevant classes and background. The fine-grained propagation module leverages the information of feature maps to propagate the visual saliency and produces high-resolution saliency maps.
Qualitative experiments show that TSGB can more discriminately explain different targets and generate clearer saliency maps than the competitive methods. 
Moreover, TSGB 
can be used
for most of the CNN models.
Quantitative experiments reveal that TSGB achieves superior localization performance, and stronger reliability over the competitive methods.
Furthermore, we also validate that TSGB is faithful to the explained models.

Note that this explanatory work is mainly based on the visual aspect, as it is difficult to establish a set of rigorous mathematical explanations. We leave the theoretical study for 
the
future research.

\ifCLASSOPTIONcaptionsoff
  \newpage
\fi

\normalem
\bibliographystyle{IEEEtran}
\bibliography{tip_bib}

\vspace{-8ex}

\begin{IEEEbiography}[{\includegraphics[width=1in,height=1.25in,clip,keepaspectratio]{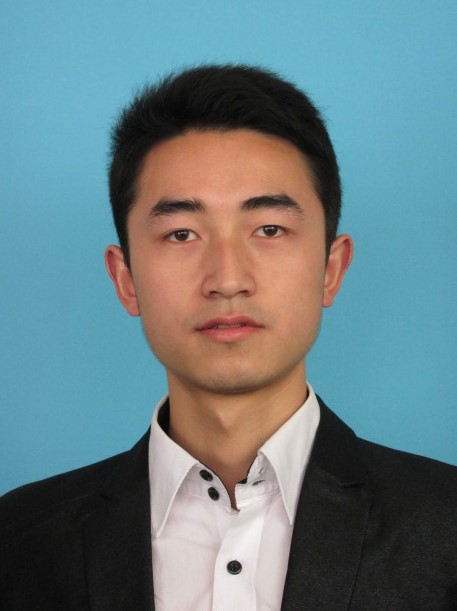}}]{Lin Cheng}
received the M.S. degree in mechanical engineering from Hangzhou Danzi University, Hangzhou, China, in 2017. He is currently pursuing the Ph.D. degree with the School of Informatics, Xiamen University, Xiamen, China. His current research interests include computer vision, machine learning, and deep learning theory. 
He serves as 
a reviewer of CVPR and ICCV.
\vspace{-10ex}
\end{IEEEbiography}

\begin{IEEEbiography}[{\includegraphics[width=1in,height=1.25in,clip,keepaspectratio]{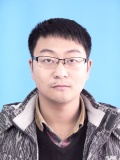}}]{Pengfei Fang} received the B.E. degree in Automation from Hangzhou Dianzi University and the M.E. degree in Mechatronics from the Australian National University (ANU), in 2014 and 2017. He is now a Ph.D. candidate at the ANU and Data61-CSIRO. He is also a visiting scholar in the Westlake University. His research interests include computer vision, machine learning, cooperative control. 
\vspace{-10ex}
\end{IEEEbiography}

\begin{IEEEbiography}[{\includegraphics[width=1in,height=1.25in,clip,keepaspectratio]{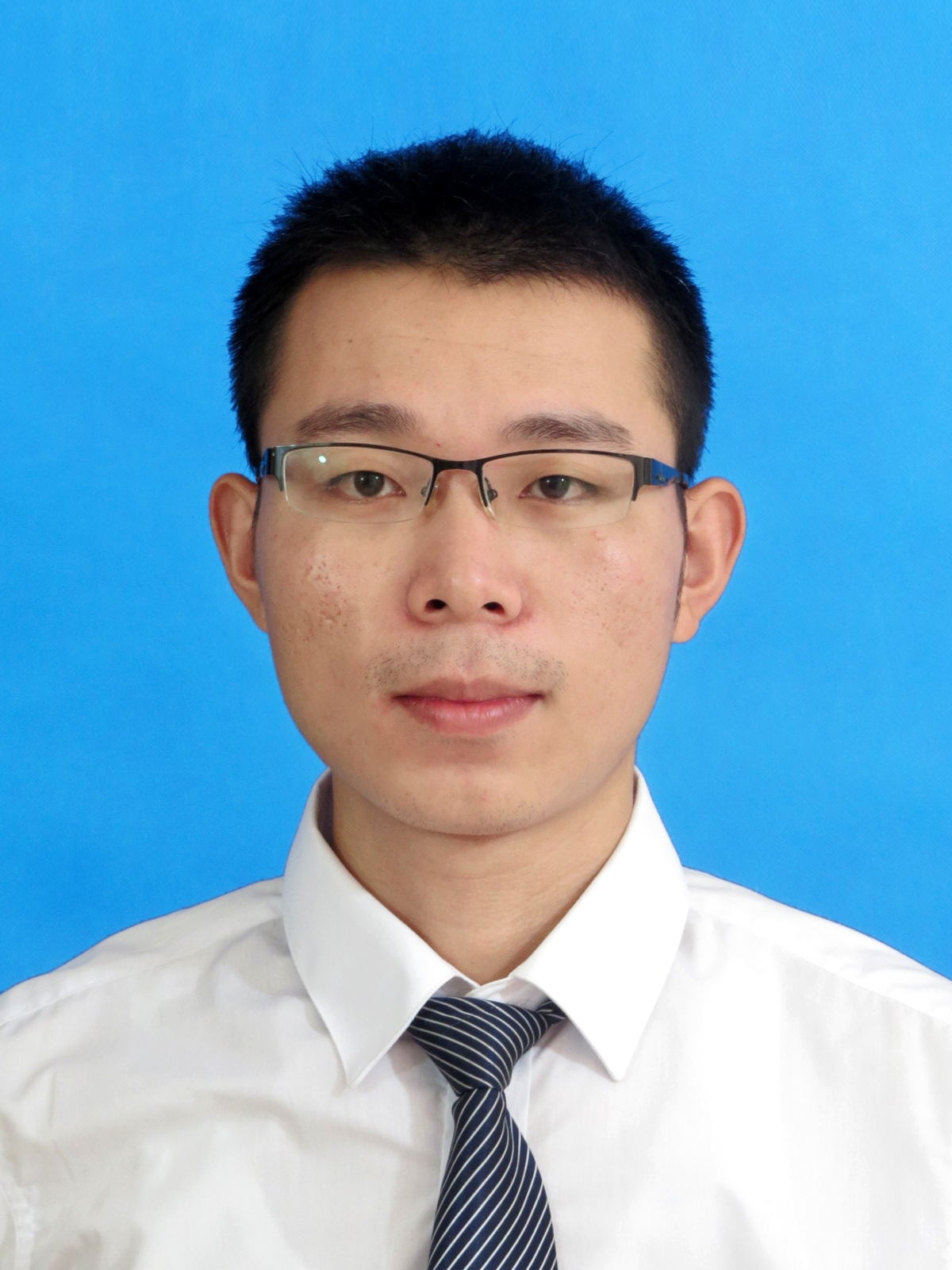}}]{Yanjie Liang}
received the Ph.D. degree with the School of Informatics, Xiamen University, Xiamen, China, in 2021. He is currently a postdoctoral researcher with the Peng Cheng Laboratory, China. He has published several papers in IEEE TITS, Pattern Recognition, ACM MM, ICME, and ACCV. His current research interests include computer vision, machine learning, and visual tracking.
\vspace{-10ex}
\end{IEEEbiography}

\begin{IEEEbiography}[{\includegraphics[width=1in,height=1.25in,clip,keepaspectratio]{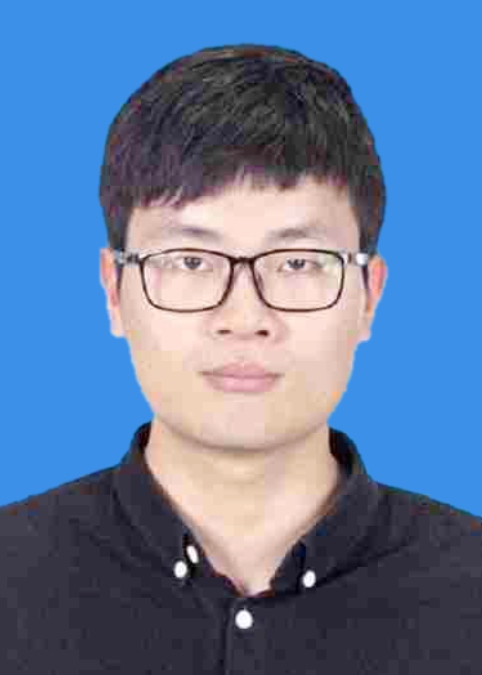}}]{Liao Zhang}
received the B.E. degree in internet of thing from Fuzhou University, Fuzhou, China, in 2016, the M.S. degree in computer science from Xiamen University, Xiamen, China, in 2020.
His research interests include computer vision and machine learning.
\vspace{-10ex}
\end{IEEEbiography}

\begin{IEEEbiographynophoto}{Chunhua Shen}
is a Professor of Computer Science at Zhejiang University, China.
\vspace{-10ex}
\end{IEEEbiographynophoto}

\begin{IEEEbiography}[{\includegraphics[width=1in,height=1.25in,clip,keepaspectratio]{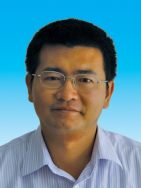}}]{Hanzi Wang}
is currently a Distinguished Professor of Minjiang Scholars in Fujian province and a Founding Director of the Center for Pattern Analysis and Machine Intelligence (CPAMI) at XMU. He was an Adjunct Professor (2010-2012) and a Senior Research Fellow (2008-2010) at the University of Adelaide, Australia; an Assistant Research
Scientist (2007-2008) and a Postdoctoral Fellow (2006-2007) at the Johns Hopkins University; and a Research Fellow at Monash University, Australia (2004-2006). He received his Ph.D. degree in Computer Vision from Monash University where he was awarded the Douglas Lampard Electrical Engineering Research Prize and Medal for his Ph.D. thesis. His research interests are concentrated on computer vision and pattern recognition. He was an Associate Editor for IEEE Transactions on Circuits and Systems for Video Technology (2010-2015) and he was a Guest Editor of Pattern Recognition Letters (September 2009). He has also served on the program committee (PC) of ICCV, ECCV, CVPR, IJCAI, SIGGRAPH, AAAI, etc, and he has served on the reviewer panel for more than 40 journals and conferences. 
\end{IEEEbiography}

\end{document}